\documentclass[journal, final, twocolumn]{IEEEtran}
\ifCLASSINFOpdf
\else
\fi
\usepackage{amsmath}
\usepackage{graphicx}
\usepackage{amssymb}
\usepackage[cmintegrals]{newtxmath}
\usepackage[english]{babel}
\usepackage[utf8x]{inputenc}
\usepackage[T1]{fontenc}

\usepackage{colortbl}
\usepackage[font=scriptsize]{caption}
\usepackage[font=footnotesize]{subcaption}
\usepackage[colorinlistoftodos]{todonotes}
\usepackage[colorlinks=true, allcolors=blue]{hyperref}
\usepackage[keeplastbox]{flushend}
\usepackage{acronym}
\usepackage{dsfont}
\usepackage{color}
\usepackage{multirow}
\usepackage{lastpage,algorithm,algorithmic}
\usepackage{epsfig,amssymb}
\usepackage{bbding}
\usepackage{booktabs}
\usepackage{color, balance}
\usepackage{cite}
\usepackage{enumitem}
\usepackage{wrapfig}
\usepackage{picins}
\usepackage{cutwin}

\newtheorem{remk}{\textbf{Remark}}

\hypersetup{final}
\begin{document}
%
\title{Resource Management in Wireless Networks via \\ Multi-Agent Deep Reinforcement Learning}
%
%
%

\author{Navid Naderializadeh, Jaroslaw Sydir, Meryem Simsek, and Hosein Nikopour
\thanks{A short version of this paper has been presented at the 21\textsuperscript{st} IEEE International Workshop on Signal Processing Advances in Wireless Communications (SPAWC 2020)~\cite{spawc2020_DRL_RRM}.

N. Naderializadeh is with HRL Laboratories, LLC, Malibu, CA, USA. J. Sydir and H. Nikopour are with Intel Corporation, Santa Clara, CA, USA. M. Simsek is with VMware, Inc., Palo Alto, CA, USA. E-mails: nnaderializadeh@hrl.com, $\{$jerry.sydir, hosein.nikopour$\}$@intel.com, msimsek@vmware.com. This work was done while N. Naderializadeh and M. Simsek were at Intel Corporation.
}}

\maketitle

\begin{abstract}
We propose a mechanism for distributed resource management and interference mitigation in wireless networks using multi-agent deep reinforcement learning (RL). We equip each transmitter in the network with a deep RL agent that receives delayed observations from its associated users, while also exchanging observations with its neighboring agents, and decides on which user to serve and what transmit power to use at each scheduling interval. Our proposed framework enables agents to make decisions simultaneously and in a distributed manner, unaware of the concurrent decisions of other agents. Moreover, our design of the agents' observation and action spaces is scalable, in the sense that an agent trained on a scenario with a specific number of transmitters and users can be applied to scenarios with different numbers of transmitters and/or users. Simulation results demonstrate the superiority of our proposed approach compared to decentralized baselines in terms of the tradeoff between average and 5\textsuperscript{th} percentile user rates, while achieving performance close to, and even in certain cases outperforming, that of a centralized information-theoretic baseline. We also show that our trained agents are robust and maintain their performance gains when experiencing mismatches between train and test deployments.
\end{abstract}


%
\IEEEpeerreviewmaketitle

\section{Introduction} 

One of the key drivers for improving throughput in future wireless networks, including fifth and sixth generation mobile networks (5G and 6G), is the densification achieved by deploying more base stations~\cite{saad2020sixgwireless}. The rise of such \emph{ultra-dense} network paradigms implies that the limited physical wireless resources (in time, frequency, etc.) need to support an increasing number of simultaneous transmissions. Effective radio resource management procedures are, therefore, critical to mitigate the interference among such concurrent transmissions and achieve the desired performance enhancement in these ultra-dense environments.

The radio resource management problem is in general non-convex and therefore computationally complex, especially as the network size increases. There is a rich literature of centralized and distributed algorithms for radio resource management, using various techniques in different areas such as geometric programming~\cite{gjendemsjo2008binary}, weighted minimum mean square optimization~\cite{shi2011iteratively}, game theory~\cite{song2014game}, information theory~\cite{naderializadeh2014itlinq,yi2015itlinq+}, and fractional programming~\cite{shen2017fplinq}.

Due to the dynamic nature of wireless networks, these radio resource management algorithms may, however, fail to guarantee a reasonable level of performance across all ranges of scenarios. Such dynamics may better be handled by algorithms that \emph{learn} from interactions with the environment. Particularly, frameworks that base their decision making process on the massive amounts of data that are already available in wireless communication networks are well suited to cope with these challenges. Comprehensive tutorials on the usage of machine learning (ML) techniques across a broad spectrum of wireless network problems can be found in~\cite{chen2017machine,chen2019mlforwireless}.

A specific subset of ML algorithms, called \emph{reinforcement learning (RL)} methods, are uniquely positioned in this regard. In the simplest form, RL algorithms consider an \emph{agent} that interacts with an \emph{environment} over time by receiving \emph{observations}, taking \emph{actions}, and collecting \emph{rewards}, while the environment transitions to the subsequent step, emitting a new set of observations. Over the course of these interactions, the goal of these algorithms is to train the agent to take actions that maximize its reward over time. Recent years have seen the rise of \emph{deep} RL, where deep neural networks (DNNs) are used as function approximators to estimate the probability of taking each action given each observation, and/or the \emph{value} of each observation-action pair. Deep RL algorithms have achieved resounding success in solving challenging sequential decision making problems, especially in various gaming environments, such as Atari 2600 and Go~\cite{mnih2015human,maddison2014move,silver2016mastering,berner2019dota}.

In this paper, we consider the application of deep RL techniques to the problem of \emph{distributed} user scheduling and downlink power control in multi-cell wireless networks, and we propose a mechanism for scheduling transmissions using \emph{multi-agent} deep RL (MARL) so as to be fair among all users throughout the network. We evaluate our proposed algorithm using a system-level simulator and compare its performance against several decentralized and centralized baseline algorithms. We show that our trained agents outperform two decentralized baseline scheduling algorithms in terms of the tradeoff between sum-rate (representative of ``cell-center'' users, i.e., the ones with relatively good channel conditions) and 5\textsuperscript{th} percentile rate (representative of ``cell-edge'' users with poor channel conditions). Moreover, our agents attain competitive performance as compared to a \emph{centralized} binary power control method, called ITLinQ~\cite{naderializadeh2014itlinq, naderializadeh2017ultra}.

Our proposed design for the MARL agents is scalable, and ensures that their DNN structure does not vary with the actual size of the wireless network, i.e., number of transmitters and users. We test the robustness of our trained agents with respect to changes in the environment, and demonstrate that our agents maintain their performance gains throughout a range of network configurations. We also shed light on the interpretability aspect of the agents and analyze their decision making criteria in various network conditions.

We make the following main contributions in this paper:

\begin{itemize}
\item We introduce a MARL agent, which performs joint optimization of user selection and power control decisions in a wireless environment with multiple transmitters and multiple users per transmitter.

\item We demonstrate that MARL techniques are effective even when the observations available to the agents are undersampled and delayed, due to real-world measurement feedback periods, as well as communication and processing delays.

\item We introduce a scalable design of observation and action spaces, allowing a MARL agent with a fixed-size neural network to operate in a variety of wireless network sizes and densities in terms of the total number of transmitters and/or users in the network.

\item We propose a novel method for normalizing the observation variables, which are input to the agent's neural network, using a \emph{percentile-based} pre-processing technique on an offline collected dataset from the actual train/test deployment.

\item We use a configurable reward to achieve the right balance between average and 5\textsuperscript{th} percentile user rates, representing ``cell-center'' and ``cell-edge'' user experiences, respectively.
\end{itemize}

\subsection{Related Work}
The inherent complexity of resource management in wireless networks, together with the promising resulting attained by (deep) RL techniques, has motivated a plethora of research over the past few years targeting different aspects of radio resource management using (deep) RL approaches. Among the earlier works in this domain, the work in~\cite{bennis2013self} addresses self-organizing networks (SONs) in a femto-cell/macro-cell scenario using classical RL in a game-theoretic framework to select transmit power levels for femto base stations (FBSs). The authors show that the FBSs converge to an optimal equilibrium point even when they have access only to information about their own performance. More recently, in~\cite{chen2020joint}, the authors perform joint optimization of wireless resources on the base station and transmit power on the UE to improve packet error rates in the context of federated learning over wireless networks. In~\cite{li2018intelligent, tondwalkar2019deep}, the authors use deep RL for the problem of spectrum sharing and resource allocation in cognitive radio networks. The authors in~\cite{liang2019spectrum} propose a deep MARL approach for spectrum sharing in vehicular networks, where each vehicle-to-vehicle (V2V) link acts as an agent, learning how to reuse the resources utilized by vehicle-to-infrastructure (V2I) links in order to improve system-wide performance metrics. Moreover, in~\cite{hua2019gan}, deep RL and generative adversarial networks (GANs) are leveraged to address demand-aware resource allocation in network slicing.

Most related to this paper are the recent works focusing on downlink power control in cellular networks using various deep single-agent RL and MARL architectures~\cite{ghadimi2017reinforcement,meng2019power,ahmed2019deep,nasir2019multi,zhao2019joint}. In~\cite{ghadimi2017reinforcement}, the authors consider cellular networks (of up to three cells in their experiments), where users in each cell are given an equal share of the spectrum, and power control and rate adaptation is done using deep MARL to optimize a network-wide utility function. In~\cite{meng2019power}, deep Q-learning is used to address the problem of power control for sum-rate maximization in multi-cell networks, where each agent has access to the instantaneous interference levels caused by adjacent cells at its own users. The authors in~\cite{ahmed2019deep} use a centralized deep Q-learning approach for downlink power control, where the agent has access to the global network state and makes power control decisions for all transmitters and on all frequency sub-bands. A scalable deep MARL approach for downlink power control has been proposed in~\cite{nasir2019multi}, where each transmitter exchanges instantaneous observations with its nearby transmitters at each scheduling interval before making power control decisions to maximize the network weighted sum-rate. Moreover, a centralized power control and channel allocation method using deep RL is introduced in~\cite{zhao2019joint}, where a central controller acts as a deep RL agent by observing the location information and rate requirements of all users, together with the set of transmit powers and channel allocations of all transmitters, and makes power control and channel allocation decisions to maximize the network sum-throughput.

There are, however, several drawbacks on the aforementioned prior works. First, most of these works intend to optimize a single metric or objective function, a prominent example of which is the sum-throughput of the links/users across the network. However, resource allocation solutions that optimize sum-throughput often allocate resources \emph{unfairly} among users, as they only focus on the average performance and fail to guarantee a minimum performance. Second, measurements of channels and other metrics at each node in a real-world wireless network are reported to other nodes with certain amounts of delay (either through the feedback channels or backhaul mechanisms), while many of the past works assume ideal message passing among transmitter/user nodes. Moreover, many of the aforementioned RL-based solutions are not scalable, in the sense that they do not address the possible mismatch between training and testing environments and do not consider the applicability and robustness of their solution to environment variations.

In this work, we leverage MARL to address the practical aspects of wireless resource management, especially related to feedback undersampling and latency. We are motivated by the fact that future wireless network paradigms need the deployment of intelligent access points, each of which may need to support a different number of users based on local load conditions. This implies that the access points must make resource management decisions in a distributed and dynamic way, and be robust to environment imperfections in terms of the information they receive from their associated users, as well as network variations in terms of the total number of access points and users existing across the network. Our proposed approach uses centralized training and distributed execution and trains the agents so that they perform well in a variety of corner cases and scenarios, hence distinguishing it from the prior work in the literature.

\section{System Model and Problem Formulation}\label{sec:sys_model}

\begin{figure}[t]
\centering
\includegraphics[trim = 2.7in 2.7in 2.95in 3.1in, clip,width=\linewidth]{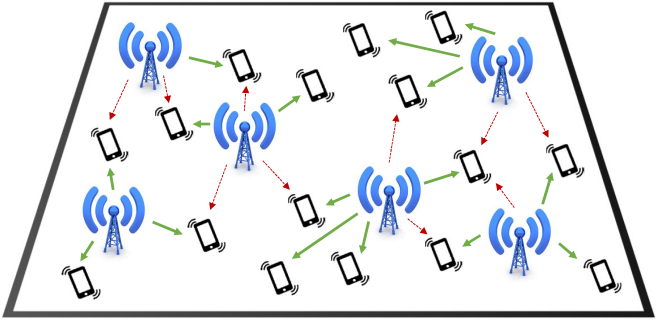}
\caption{\label{fig:sys_model}A wireless network comprising multiple access points (APs) and user equipment devices (UEs). The solid green lines denote the signal links between APs and their associated UEs, while the dashed red lines denote (strong) interference links between APs and adjacent non-associated UEs.}
\vspace{-.1in}
\end{figure}

We consider the downlink direction of a synchronous time-slotted wireless network, consisting of $N$ access points (APs) $\{\mathsf{AP}_i\}_{i=1}^N$ and $K$ user equipment devices (UEs) $\{\mathsf{UE}_j\}_{j=1}^K$, as illustrated in Figure~\ref{fig:sys_model}, where the APs intend to transmit data to the UEs across the network over a set of $T$ consecutive scheduling intervals. At each scheduling interval $t\in\{1,2,\dots,T\}$, we denote the channel gain between $\mathsf{AP}_i$ and $\mathsf{UE}_j$ as $h_{ji}(t)$, which can be decomposed as $h_{ji}(t)=H_{ji} \tilde{h}_{ji}(t)$, where $H_{ji}$ and $\tilde{h}_{ji}(t)$ respectively denote the constant long-term and time-varying short-term components of the channel gain between $\mathsf{AP}_i$ and $\mathsf{UE}_j$.

We assume that each AP maintains a fixed local pool of users associated with it during the period of $T$ scheduling intervals, which can be refreshed periodically due to user mobility, etc. User association may be performed using any arbitrary method, e.g., max-RSRP (reference signal received power)~\cite{dhillon2013load}. We denote the set of UEs associated with $\mathsf{AP}_i$ by $\mathcal{L}_i$. We assume that each AP has at least one UE associated with it and that every UE is associated with one and only one AP, ensuring that the set of user pools for all APs partitions the entire set of UEs; i.e.,
\begin{align}
\mathcal{L}_i &\neq \emptyset, ~ \forall i\in\{1,\dots,N\},\label{eq:user_pool1}\\
\mathcal{L}_i \cap \mathcal{L}_{i'} &= \emptyset, ~ \forall (i,i')\in\{1,\dots,N\}^2 \text{ s.t. } i\neq i',\\
\bigcup_{i\in\{1,\dots,N\}} \mathcal{L}_i &= \{\mathsf{UE}_1,\dots,\mathsf{UE}_K\}.\label{eq:user_pool3}
\end{align}

We focus on two major decisions that each $\mathsf{AP}_i$ makes at each scheduling interval $t$:
\begin{itemize}
\item \textbf{User scheduling}: It chooses one of the UEs from its local user pool to serve. We let $\mathsf{UE}_{j_i(t)} \in \mathcal{L}_i$ denote the UE that is scheduled to be served by $\mathsf{AP}_i$ at scheduling interval $t$.
\item \textbf{Power control}: It selects a transmit power level $P_i(t)\in[0,P_{\mathsf{max}}]$ to serve $\mathsf{UE}_{j_i(t)}$.
\end{itemize}

Given the above decisions, the received signal at $\mathsf{UE}_{j_i}$ (we removed the dependence on $t$ for brevity) at scheduling interval $t$ can be written as
\begin{align}
\vspace{-.45in}
y_{j_i}(t) = h_{j_i i}(t) x_i(t) + \sum_{k\neq i} h_{j_i k}(t) x_k(t) + n_{j_i}(t),
\end{align}
where $x_i(t)$ denotes the signal transmitted by $\mathsf{AP}_i$ at scheduling interval $t$ and $n_j(t) \sim \mathcal{CN}(0,\sigma^2)$ denotes the additive white Gaussian noise at $\mathsf{UE}_j$ in scheduling interval $t$, with $\sigma^2$ denoting the noise variance. This implies that the achievable rate of $\mathsf{UE}_{j_i}$ at scheduling interval $t$ (taken as the Shannon capacity) can be written as
\vspace{-.05in}
\begin{align}\label{eq:simple_rate}
R_{j_i}(t) = \log_2\left(1 + \frac{|h_{j_i i}(t)|^2 P_i(t)}{\sum_{k\neq i} |h_{j_i k}(t)|^2 P_k(t) + \sigma^2}\right).
\end{align}
Having~\eqref{eq:simple_rate}, we define the average rate of each $\mathsf{UE}_j$ over the course of $T$ scheduling intervals as
\begin{align}
\bar{R}_j = \frac{1}{T} \sum_{t=1}^{T} R_j(t).
\end{align}

We now specify the following two metrics that we intend to optimize in this paper:
\begin{itemize}[leftmargin=*]
\item \textbf{Sum-rate}: This metric is defined as the aggregate average throughput across the entire network over the course of $T$ scheduling intervals; i.e.,
\vspace{-.05in}
\begin{align}
R_{\mathsf{sum}} = \sum_{j=1}^K \bar{R}_j.
\end{align}

\item \textbf{5\textsuperscript{th} percentile rate}: As the name suggests, this metric is defined as the average rate threshold achieved by at least $95\%$ of the UEs over $T$ scheduling intervals. In a probabilistic sense,
\vspace{-0.05in}
\begin{align}\label{eq:5th_percentile}
R_{5\%} = \max R \text{ s.t. } \mathbb{P}[\bar{R}_j \geq R]\geq 0.95, \forall j\in\{1,...,K\}.
\end{align}
\end{itemize}

\begin{figure*}[t]
\centering
\setlength{\belowcaptionskip}{-15pt}
\includegraphics[trim = .9in 1.4in 1.2in 2.05in, clip,width=0.8\textwidth]{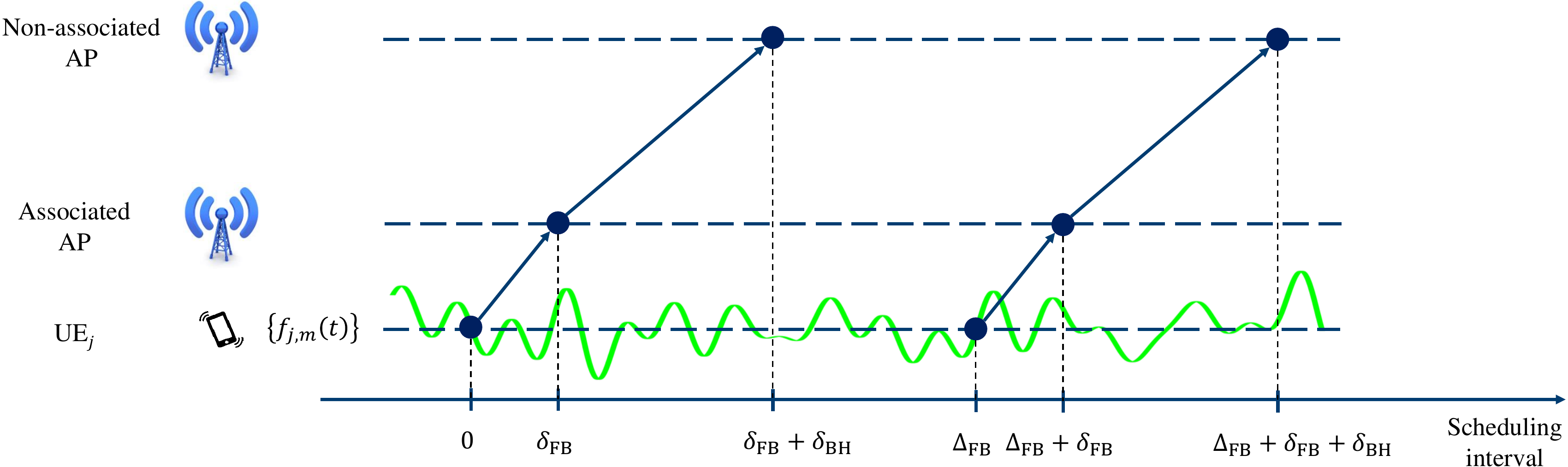}
\caption{An illustration of the timeline for reporting the status updates from the UEs back to their associated APs with a period of $\Delta_{\mathsf{FB}}$ scheduling intervals and delay of $\delta_{\mathsf{FB}}$ scheduling intervals. These feedback reports are exchanged between the APs via a backhaul interface with a further delay of $\delta_{\mathsf{BH}}$ scheduling intervals.}
\label{fig:obs_delays}
\end{figure*}

Our goal is to devise an algorithm that outputs a sequence of joint user scheduling and power control decisions across the network leading to the best trade-off between the two aforementioned metrics. Specifically, we aim to solve the following multi-objective optimization problem,
\begin{subequations}\label{eq:opt_RRM}
\begin{alignat}{2}
&\hspace{-.1in}\max_{\big\{j_i(t), P_i(t)\big\}_{(i,t)\in[N \times T]}}&&  \hspace{-.03in}\left(R_{\mathsf{sum}}, R_{5\%}\right) \\
&\quad\quad\quad\text{s.t.} &&  \eqref{eq:simple_rate}-\eqref{eq:5th_percentile}, \\
& &&  j_i(t)\in\mathcal{L}_i , ~\forall (i,t)\in[N \times T], \\
& &&  P_i(t)\in[0, P_{\mathsf{max}}] , ~\forall (i,t)\in[N \times T],
\end{alignat}
\end{subequations}
where $[N \times T]\coloneqq \{1,\dots,N\} \times \{1,\dots,T\}$. This optimization problem seeks to find the Pareto front, i.e., the set of Pareto optimal solutions that are not dominated by any other solution, where each solution is defined as a set of consecutive user selection and power control decisions. To be precise, a solution $\{j_i(t), P_i(t)\}_{(i,t)\in[N \times T]}$ is said to dominate another solution $\{j'_i(t), P'_i(t)\}_{(i,t)\in[N \times T]}$ if
\begin{align}
&R_{\mathsf{sum}}\Big( \{j_i(t), P_i(t)\}_{(i,t)\in[N \times T]} \Big) \nonumber\\
&\qquad\geq R_{\mathsf{sum}}\Big( \{j'_i(t), P'_i(t)\}_{(i,t)\in[N \times T]} \Big) \label{eq:dominate1} \\
&R_{5\%}\Big( \{j_i(t), P_i(t)\}_{(i,t)\in[N \times T]} \Big) \nonumber\\
&\qquad\geq R_{5\%}\Big( \{j'_i(t), P'_i(t)\}_{(i,t)\in[N \times T]} \Big),  \label{eq:dominate2}
\end{align}
with at least one of the inequalities in~\eqref{eq:dominate1}-\eqref{eq:dominate2} being strict. Note that the sum-rate indicates how high the throughputs of the users are on average across the network (i.e., ``cell-center'' users), while the 5\textsuperscript{th} percentile rate shows the performance of the worst-case users with poor channel conditions (i.e., ``cell-edge'' users), hence making these two metrics in natural conflict with each other. The problem in~\eqref{eq:opt_RRM} is challenging to solve, due to its combinatorial and non-linear structure. In Section~\ref{sec:RLframework}, we show how we can leverage deep MARL to tackle this problem.\footnote{Note that as opposed to tabular RL algorithms, the proposed deep RL algorithm may only converge to a \emph{local} optimum of the optimization problem in~\eqref{eq:opt_RRM} and global optimality is not guaranteed in general.}

\subsection{Distributed Scheduling, Feedback and Backhaul Delays}\label{sec:FB_delays}
We particularly aim to design a \emph{distributed} scheduling algorithm in the sense each AP should make its decisions on its own. To that end, the APs rely on measurements periodically made by the UEs. We assume that each $\mathsf{UE}_{j} \in \mathcal{L}_i$ measures $M$ status indicators $\{f_{j,m}(t)\}_{m=1}^M$ and reports them back to its associated $\mathsf{AP}_i$ every $\Delta_{\mathsf{FB}}$ scheduling intervals and that these feedback reports arrive at the AP with a certain delay of $\delta_{\mathsf{FB}}$ scheduling intervals. We further assume that the APs share these measurements with their neighbors via a backhaul interface with an additional delay of $\delta_{\mathsf{BH}}$ scheduling intervals. Figure~\ref{fig:obs_delays} visualizes how feedback reports are exchanged between the UEs and their associated and non-associated APs over time.\footnote{Note that the use of feedback adds to the complexity of our algorithm.}

\section{Proposed Deep MARL Framework}\label{sec:RLframework}
\begin{figure*}[t]
\centering
\setlength{\belowcaptionskip}{-15pt}
\includegraphics[trim = 1.1in 1.85in 0.94in 2.4in, clip,width=0.8\textwidth]{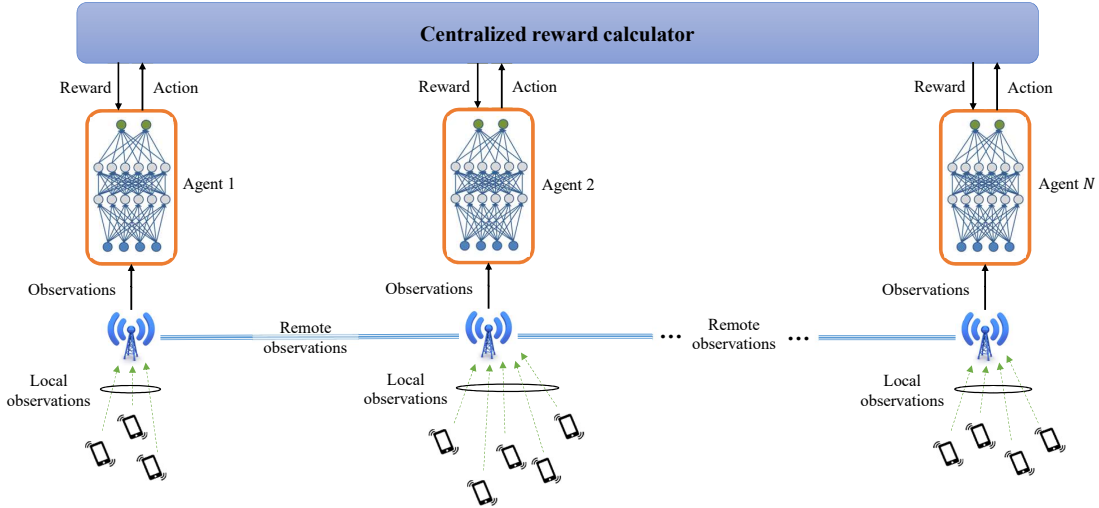}
\caption{Deep MARL diagram, where the agents receive \emph{local observations} from their associated UEs, while also receiving \emph{remote observations} from their neighboring agents. Upon taking their actions, the agents receive a centralized reward, which helps them tune their policies to take actions that maximize their rewards over time.}
\label{fig:MARL}
\end{figure*}

We use MARL to (approximately) solve the optimization problem in~\eqref{eq:opt_RRM} in a distributed manner. In particular, we propose to equip each AP with a deep MARL agent, as illustrated in Figure~\ref{fig:MARL}. As mentioned in Section~\ref{sec:FB_delays}, each agent observes the state of the UEs in its local user pool, and it also exchanges information with neighboring agents, thus observing the state of neighboring APs' associated UEs. We consider the centralized training and distributed execution paradigm, as considered in many prior works in the deep MARL literature~\cite{jorge2016learning, tampuu2017multiagent, foerster2018counterfactual, sunehag2018value, rashid2018qmix}, where each agent makes its own decision at each scheduling interval only based on the specific observations it receives from the environment. We use the experiences of all agents to train a single policy, which is used by all of the agents.

Although our overall framework is based on a standard application of deep RL techniques, such as independent Q-learning (IQL) proposed in~\cite{tampuu2017multiagent}, we define a unique approach to handling variable numbers of users and network densities using a fixed size neural network. We further describe a unique method for normalization of neural network inputs, which takes into account differences in the distribution of input parameter values for different deployment environments. Finally, our formulation takes communication and measurement delays into account, rather than assuming that agents have instantaneous access to information.

\subsection{Environment}

The environment is parametrized by the physical size of the deployment area, number of APs and UEs, and parameters governing the channel models used to create AP-UE channel realizations. At the start of each~\emph{episode}, the environment is reset, resulting in a new set of physical locations for all the APs and UEs in the network, along with channel realizations between them, modeling both long-term and short-term fading components.

\subsubsection{Observations}\label{sec:obs}
Observations available to each agent at each scheduling interval consist of \emph{local observations}, representing the state of the UEs associated with the corresponding AP and \emph{remote observations}, representing the state of the UEs associated with neighboring APs.

\begin{itemize}[leftmargin=*]
\item \textbf{Local observations}: 
We consider the case where each UE reports $M=2$ measurements to its associated AP, namely its \emph{weight} and \emph{signal-to-interference-plus-noise ratio (SINR)}. We define the weight of each $\mathsf{UE}_j, j\in\{1,...,K\}$ at scheduling interval $t$ as
\begin{align}
f_{j,1}(t) = w_j(t) =\frac{1}{\bar{R}_j(t)},
\end{align}
where $\bar{R}_j(t)$ represents the \emph{long-term average rate} of $\mathsf{UE}_j$ at scheduling interval $t$, defined as
\begin{align}
\bar{R}_j(t) &= (1-\alpha_R) \bar{R}_j(t-1) + \alpha_R R_j(t-1)\label{eq:moving_ave1} \\
&= \sum_{\tau=1}^{t-1} \alpha_R (1-\alpha_R) ^{t-\tau-1} R_j(\tau).\label{eq:moving_ave2}
\end{align}
In the above equations, $\alpha_R\in(0,1)$ is a parameter close to zero, which specifies the window size for the exponential moving average operation. Moreover, for each $\mathsf{AP}_i, i\in\{1,...,N\}$ we define the measured SINR of $\mathsf{UE}_{j}, j\in\mathcal{L}_i$ at scheduling interval $t$ as
\begin{align*}
f_{j,2}(t) = \mathsf{SINR}_j(t) = \frac{|h_{ji}(t)|^2 P_{\mathsf{max}}}{\bar{I}_j(t)+\sigma^2},
\end{align*}
where $\bar{I}_j(t)$ denotes the \emph{long-term average interference} received by $\mathsf{UE}_{j}$ up to scheduling interval $t$, and is calculated recursively as
\begin{align*}
\bar{I}_j(t) &= (1-\alpha_I) \bar{I}_j(t-1) + \alpha_I \sum_{k\neq i} |h_{jk}(t-1)|^2 P_k(t-1) \\
&= \sum_{\tau=1}^{t-1}  \sum_{k\neq i} \alpha_I (1-\alpha_I)^{t-\tau-1} |h_{jk}(\tau)|^2 P_k(\tau),
\end{align*}
with $\alpha_I\in(0,1)$ being a parameter close to zero that determines the window size for the exponential moving average of the interference.

The number of UEs associated with each AP can be different from AP to AP and from deployment to deployment. For our algorithm to be applicable to any scenario, we bound the dimension of the observation space by including observations (weights and SINRs) from a constant number of UEs per AP in any environment configuration, which we denote by $k$. To select the \emph{top-$k$} UEs, we use the \emph {proportional-fairness (PF)} ratio, defined as
\begin{align}\label{eq:PF_def}
\mathsf{PF}_j(t) = w_j(t) \cdot \log_2\left(1 + \mathsf{SINR}_j(t)\right).
\end{align}
The PF ratio provides a priority notion for the UEs, where those with higher PF ratios are more in need to be served. Hence, at each scheduling interval, each AP sorts its associated UEs based on their PF ratios, and selects the top-$k$ to include in its local observation vector. Note that the selection of the top-$k$ UEs is performed at every scheduling interval. This ensures that all UEs are serviced over time because the PF ratio of a UE that is not selected increases as its long-term average rate decreases.

\item \textbf{Remote observations}: We bound the number of agents whose observations are included in each agent's observation vector to a fixed number denoted by $n$. The \emph{top-$n$ remote agents} are the agents whose APs are physically closest, as they are likely to be the strongest interferers.

\end{itemize}

It can be verified that as there are $n$ remote agents per agent, and $k$ UEs' observations are included per agent, the length of the observation vector for each agent equals $2(n+1)k$.

\begin{remk}
Note that the dimension of the observation space, and therefore the input size to the deep RL agent's neural network, does not depend on the number of APs and/or UEs in the network. This makes our algorithm scalable regardless of the specific environment parameters.
\end{remk}

\begin{remk}
When there are fewer than $k$ UEs associated to an AP, we set the corresponding values in the local/remote observations to default values, similar to a zero-padding operation. In particular, we use default values of 0 for weight and -60 dB for SINR.
\end{remk}

\begin{remk}
The LTE and 5G standards developed by 3GPP support periodic channel quality indicator (CQI) feedback reports from UEs to their serving base stations. Moreover, the weights can either be reported by the UEs, or the base stations may keep track of the long-term average rates of their associated UEs. The base stations can also exchange observations among each other through backhaul links, such as the X2 interface~\cite{nardini2016modeling}. Therefore, our proposed observation structure is practical and may be readily implemented in current and future cellular networks.
\end{remk}

\subsubsection{Actions}\label{sec:actions}
To jointly optimize the user scheduling and power control decisions, we define a joint action space, where each action represents a (transmit power level, target UE) pair. We quantize the range of \emph{positive} transmit powers $(0, P_{\mathsf{max}}]$ to $p$ (potentially non-uniform) power levels. Moreover, as the number of UEs associated with an AP can be varying and/or potentially large, at each scheduling interval, we limit the choice of the target UE to one of the top-$k$ UEs included in the local observations. This is also reasonable because the agent has information solely on those $k$ users in its local observation vector.

Given the above considerations, the number of possible actions for each agent at each scheduling interval is $1+pk$, where the additional action is one in which the agent remains silent for that scheduling interval and selects none of the top-$k$ UEs to serve. In the event that the AP has fewer than $k$ associated UEs and erroneously selects a target UE which does not exist, we map the action to the ``off'' action, indicating that the AP should not transmit.

\begin{remk}
Note that quantizing the power levels enables us to use a wide variety of deep RL agents with discrete action spaces, and prior work has also shown that binary power control can be optimal in many cases~\cite{gjendemsjo2008binary}. Nevertheless, the framework can be extended to continuous power control using other agent architectures with continuous action spaces, such as deep deterministic policy gradient (DDPG)~\cite{lillicrap2015continuous} or soft actor-critic (SAC)~\cite{haarnoja2018soft}, which we leave for future work.
\end{remk}

\begin{remk}
Note that similar to the observation space, the action space dimension does not depend on the network size. This allows us to have a robust agent architecture that can be trained in a specific environment, and then deployed on a different environment in terms of, for instance, number of APs and/or UEs compared to the training environment. In Section~\ref{sec:cross_tests}, we show how well the agent performs in such mismatched scenarios.
\end{remk}

\begin{remk}
In this paper, we consider a single resource in the frequency domain, i.e., sub-band, at each scheduling interval. In case of multiple sub-bands, an instance of the proposed algorithm can be applied independently on each sub-band. We leave the joint optimization of user scheduling and power control across multiple sub-bands as future work.
\end{remk}

\begin{remk}
The inference complexity of the proposed algorithm over $T$ scheduling intervals scales as $\mathcal{O}\left(N^2 \log n + TK\log k + TN (n+p)k \right)$ across the entire network, where $N$ and $K$ respectively denote the total number of APs and UEs, $n$ denotes the number remote agents per AP, $k$ denotes the number of top UEs considered by every AP at each scheduling interval, and $p$ denotes the number of positive transmit power levels. The terms in the summation correspond to the complexity of initial identification of the remote agents, sorting the UEs based on PF ratios, and a forward pass of the observations through the agent neural networks, respectively, where the number of neural network parameters is treated as a constant. Note that the inference complexity per scheduling interval increases linearly with the number of APs and UEs, as opposed to the quadratic complexity of centralized scheduling algorithms, such as ITLinQ~\cite{naderializadeh2014itlinq, naderializadeh2017ultra}.
\end{remk}

\subsubsection{Rewards}
As shown in Figure~\ref{fig:MARL}, we utilize a centralized reward based on the actions of all the agents at each scheduling interval. In particular, assuming that each $\mathsf{AP}_i, i\in\{1,...,N\}$ has selected $\mathsf{UE}_{j_i}$ to serve at scheduling interval $t$, the reward emitted to each of the agents is a \emph{weighted sum-rate} reward, calculated as
\vspace{-.1in}
\begin{align}\label{eq:reward}
r(t) = \sum_i \left(w_{j_i}^i(t)\right)^{\lambda_{\mathsf{rew}}}  R_{j_i}(t),
\end{align}
where $w_{j_i}^i(t)$ is the most recent reported weight measurement by $\mathsf{UE}_{j_i}$ available at $\mathsf{AP}_i$, $R_{j_i}(t)$ is the rate achieved by $\mathsf{UE}_{j_i}$, and $\lambda_{\mathsf{rew}}\in[0,1]$ is a parameter which determines the tradeoff between $R_{\mathsf{sum}}$ and $R_{5\%}$, the two metrics that we intend to optimize. Specifically, $\lambda_{\mathsf{rew}}=0$ turns the reward to sum-rate, favoring cell-center UEs, while $\lambda_{\mathsf{rew}}=1$ changes the reward to approximately the summation of the scheduled UEs' PF ratios, hence appealing to cell-edge UEs.

There are, however, two exceptions to the reward emitted to the agents, which are as follows:
\begin{itemize}
\item \textbf{All agents off}: Due to the distributed nature of decision making by the agents, it is possible that at a scheduling interval, all agents decide to remain silent. This is clearly a suboptimal joint action vector at any scheduling interval. Therefore, in this case, we penalize the agent, whose top user has the highest PF ratio among all UEs in the network, with the negative of that PF ratio, while the rest of the agents will receive a zero reward, according to~\eqref{eq:reward}.

\item \textbf{Invalid user selected}: As mentioned in Section~\ref{sec:actions}, it might happen that an AP has fewer than $k$ associated UEs in its user pool, and selects an invalid UE to serve at a scheduling interval. In that case, the agent corresponding to that AP is penalized by receiving a zero reward regardless of the actual weighted sum-rate reward of the other agents as given in~\eqref{eq:reward}.
\end{itemize}

\vspace{-.2in}
\subsection{Normalizing Agents' Observations and Rewards}
It is widely known that normalizing DNN inputs and outputs significantly impacts its training behavior and predictive performance~\cite{lecun2012efficient}. Before training or testing our model in a specific environment, we first create a number of offline environment realizations and run one or several simple baseline scheduling algorithms in those realizations. While doing so, we collect data on the observations and rewards of all the agents throughout all environment realizations and all scheduling intervals within each realization. 
We then leverage the resulting dataset as follows to pre-process the observations and rewards before using them to train the agents:
\begin{itemize}[leftmargin=*]
\item \textbf{Observations:} As mentioned in Section~\ref{sec:obs}, we have two types of observations: weights and SINRs. For notational simplicity, we describe the normalization process for the weight observations; the process for normalizing SINR observations follows similarly.

\begin{figure*}[t]
\centering
\setlength{\belowcaptionskip}{-10pt}
\includegraphics[width=.48\textwidth]{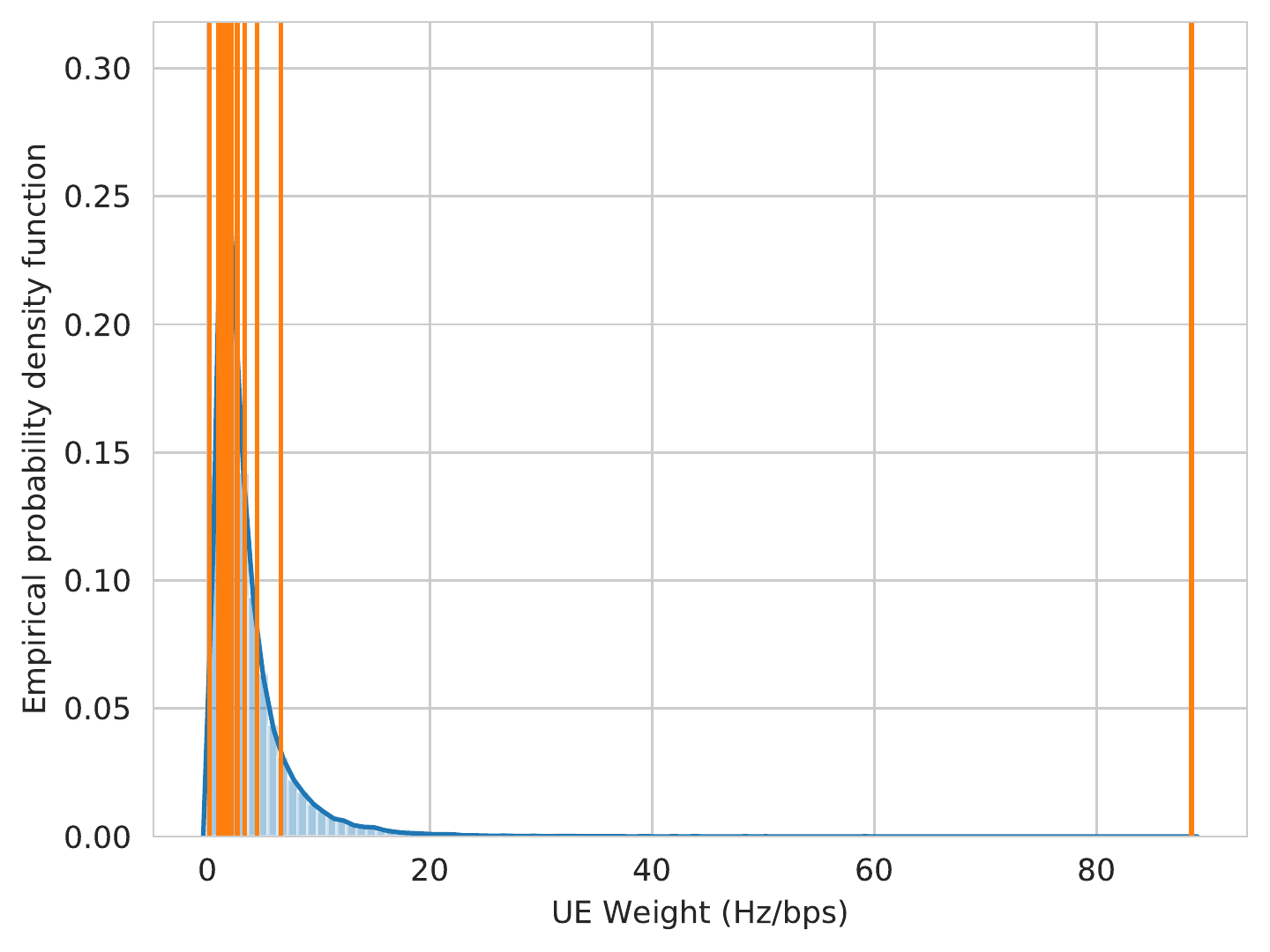}
~
\includegraphics[width=.48\textwidth]{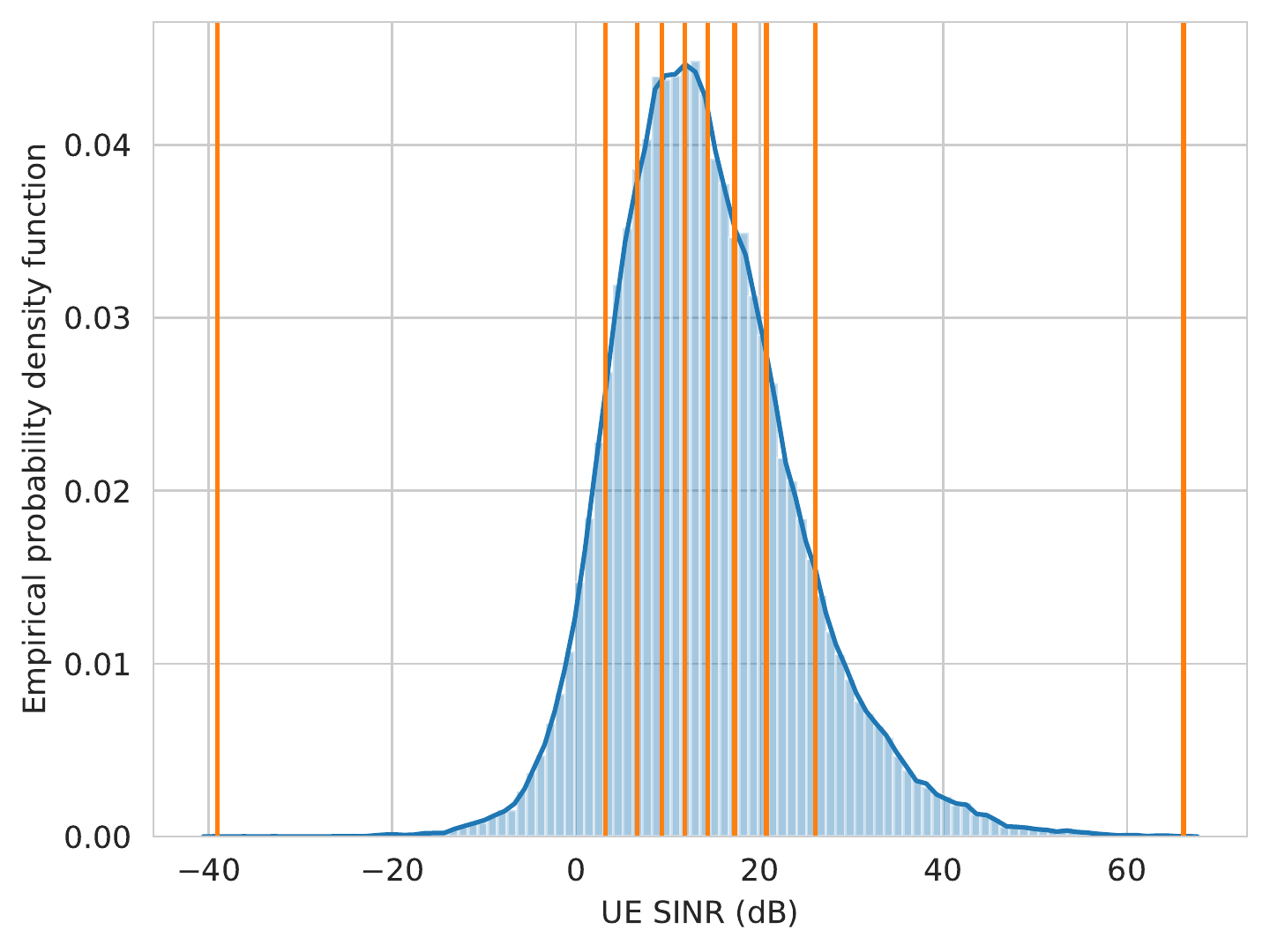}
\caption{Example empirical distributions of weight (left) and SINR (right) observations (in blue) and their corresponding percentiles (in orange) using $Q=10$ percentile levels.}
\label{fig:percentiles}
\end{figure*}

Considering the weight observations, we derive the empirical distribution of the observed weights in the aforementioned dataset. We then use the distribution to calculate multiple \emph{percentiles} of the observed weights. In particular, we consider $Q$ percentiles at \linebreak $\left\{0,\frac{100}{Q-1}, \frac{200}{Q-1}, ..., 100\right\}\%$, denoted respectively by $\left\{p_{w,0},p_{w,1},p_{w,2}, ..., p_{w,Q-1}\right\}$, as depicted in Figure~\ref{fig:percentiles} (for both weights and SINRs). Note that $p_{w,0}$ and $p_{w,Q-1}$ are equal to the minimum and maximum weights observed in the dataset, respectively. Afterwards, we map each subsequent weight observation $w$ during training/inference before feeding to the neural network as
\begin{align}\label{eq:obs_mapping}
\hat{w} =
\begin{cases}
-\frac{1}{2} &\text{ if } w < p_{w,0},\\
\frac{q+1}{Q}-\frac{1}{2} &\text{ if } p_{w,q} \leq w < p_{w,q+1}, q\in\{0,...,Q-2\}, \\
\frac{1}{2} &\text{ if } w \geq p_{w,Q-1}.
\end{cases}
\end{align}
The mapping in $\eqref{eq:obs_mapping}$ is in fact applying a (shifted version of the) CDF of the weight observation to itself, which is known to be uniformly distributed. Therefore, this mapping guarantees that the observations fed into the neural network will (approximately) follow a discrete uniform distribution over the set $\left\{-\frac{1}{2}, \frac{1}{Q}-\frac{1}{2} , \frac{2}{Q}-\frac{1}{2}, ..., \frac{Q-1}{Q}-\frac{1}{2}, \frac{1}{2} \right\}$.

\item \textbf{Rewards:} We follow a well-known standardization procedure for normalizing the rewards, where we use the dataset to estimate the mean and standard deviation of the reward, denoted by $\mu_{\mathsf{rew}}$ and $\sigma_{\mathsf{rew}}$, respectively. Each reward during training is then normalized as
\begin{align}
\hat{r} = \frac{r - \mu_{\mathsf{rew}}}{\sigma_{\mathsf{rew}}},
\end{align}
ensuring that the neural network outputs have (approximately) zero mean and unit variance.
\end{itemize}

\begin{table*}[!thb]
\caption{Simulation parameters.}\label{table:params}
\begin{subtable}{.38\linewidth}
\caption[fontsize=tiny]{Wireless environment}
\centering
\footnotesize
\noindent\makebox[.8\textwidth]{
\setlength\tabcolsep{6pt} 
\setlength{\aboverulesep}{0pt}
\setlength{\belowrulesep}{0pt}
\resizebox{\textwidth}{!}{
\begin{tabular}{l|l}
\cmidrule[1.5pt]{1-2}
Parameter & Value \\ \hline
Number of APs ($N$) & $4-10$ \\
Number of UEs ($K$) & $16-40$ \\
Network area & 500m$\times$500m \\
Minimum AP-AP distance & 35m \\
Minimum AP-UE distance & 10m \\
Bandwidth & 10MHz \\
Maximum transmit power ($P_{\mathsf{max}}$) & 10dBm \\
Noise PSD & -174dBm/Hz \\
Episode length ($T$) & 2000 \\
Feedback report period ($\Delta_{\mathsf{FB}}$) & 10 \\
Feedback report delay ($\delta_{\mathsf{FB}}$) & 5 \\
Backhaul delay ($\Delta_{\mathsf{BH}}$) & 5 \\
Rate averaging parameter ($\alpha_R$) & 0.01 \\
Interference averaging parameter ($\alpha_I$) & 0.05 \\
Number of observable UEs per agent ($k$) & 3 \\
Number of remote APs per agent ($n$) & 3 \\
Number of positive power levels ($p$) & 1 \\
Number of percentile levels ($Q$) & 20 \\
Reward weight exponent ($\lambda_{\mathsf{rew}}$) & 0.8 \\
Reward discount factor ($\gamma$) & 0.9 \\
\cmidrule[1.5pt]{1-2}
\end{tabular}}}
\end{subtable}
~~~~~~
\begin{subtable}{.595\linewidth}
\caption{Deep MARL agents}\label{table:agent_params}
\centering
\footnotesize
\noindent\makebox[.8\textwidth]{
\setlength\tabcolsep{6pt} 
\setlength{\aboverulesep}{0pt}
\setlength{\belowrulesep}{0pt}
\resizebox{\textwidth}{!}{
\begin{tabular}{cl|lc}
\cmidrule[1.5pt]{2-3}
& Parameter & Value \\ \cline{2-3}
& Number of parallel environments & $4$ \\
& Experience buffer size & $25,000$ \\
& Optimizer & \texttt{Adam} \\
& Batch size & $1024N$ \\
& Target network update period & $10,000$ intervals \\
& Learning rate (schedule) & $0.01$ ($\times \frac{1}{2}$ every 5,000 iterations) \\
\multirow{-7}{*}{\rotatebox[origin=c]{90}{DQN}} & Policy & $\epsilon$-greedy ($\epsilon: 100\% \rightarrow 1\%$ over $25$ episodes) & \multirow{-7}{*}{\rotatebox[origin=c]{90}{\phantom{DQN}}} \\ \cline{2-3}
& Number of parallel environments & $10$ \\
& Optimizer & \texttt{RMSProp} ($\epsilon=10^{-5}, \alpha=0.99$) \\
& Batch size & $1000N$ \\
& Learning rate (schedule) & $5\times10^{-4}$ ($\times \frac{1}{2}$ every 12,000 iterations) \\
& Gradient clipping threshold & $1$ \\
& Policy loss coefficient & $1$ \\
& Value function loss coefficient & $1$ \\
\multirow{-8}{*}{\rotatebox[origin=c]{90}{A2C}} & Entropy regularization coefficient & $0.05$ \\ \cline{2-3}
& Number of layers & $2$ \\
& Number of neurons per layer & $128$ \\
& Activation function & \texttt{tanh} \\
& Training period & $100$ intervals \\
\multirow{-6}{*}{\rotatebox[origin=c]{90}{Shared}} & $L_2$ regularization coefficient & $0.001$ \\
\cmidrule[1.5pt]{2-3}
\end{tabular}}}
\end{subtable}%
\end{table*}
\vspace{-.1in}

\subsection{Training and Validation Procedure}\label{sec:validation_proc}
We consider an episodic training procedure, where in each episode,
the locations of the APs and UEs and channel realizations are randomly selected following a set of probability distributions and constraints on minimum AP-AP and UE-AP distances. We control the density of APs and UEs in our environment by fixing the size of the deployment area and selecting different numbers of APs and UEs for different training sessions. Because the channels between APs and UEs depend heavily on their relative locations, each new episode allows the system to experience a potentially unexplored subset of the observation space. The UE-AP associations take place based on~\eqref{eq:def_UEpool} as a new episode begins, and remain fixed for the duration of that episode. An episode consists of a fixed number of $T$ scheduling intervals, where at each interval, the agents decide on which user scheduling and power control actions the APs should take.

\section{Simulation Results}\label{sec:results}

In this section, we first describe the wireless channel model that we use in our experiments. We then present the baseline algorithms that we use to compare the performance of our proposed method against. Next, we discuss our considered deep MARL agents. Afterwards, we proceed to present our simulation results. Finally, we dig deeper into interpreting the agents' learned decision making criteria. The simulation parameters are shown in Table~\ref{table:params}.

\subsection{Wireless Channel Model}
The communication channel between APs and UEs consists of three different components:
\begin{itemize}
\item \textbf{Path-loss}: We consider a dual-slope path-loss model~\cite{andrews2016we,zhang2015downlink}, which states that the path-loss at distance $d$ equals
\begin{align*}
\mathsf{PL}(d)=\begin{cases}
K_0 d^{\alpha_1} &\text{if }d \leq d_{\mathsf{BP}}\\
K_0 \frac{d^{\alpha_2}}{d_{\mathsf{BP}}^{\alpha_2-\alpha_1}} &\text{if }d > d_\mathsf{BP}
\end{cases},
\end{align*}
where $K_0$ denotes the path-loss at distance $d=1$m, $d_{\mathsf{BP}}$ denotes the break-point distance, and $\alpha_1$ and $\alpha_2$ denote the path-loss exponents before and after the break-point distance, respectively ($\alpha_1\leq\alpha_2$). In this paper, we set $K_0=39$ dB, $\alpha_1=2$, $\alpha_2=4$, and $d_{\mathsf{BP}}=100$m. 

\item \textbf{Shadowing}: All the links experience log-normal shadowing with a $7$dB standard deviation.

\item \textbf{Short-term fading}: We use the sum of sinusoids (SoS) model~\cite{li2002simulation} for short-term flat Rayleigh fading (with pedestrian node velocity of $1$m/s) in order to model the dynamics of the communication channel over time.
\end{itemize}

\subsection{User Association}
We use the \emph{max-RSRP} (reference signal received power) user association policy, where each UE is associated to the AP from which it receives the highest average power over time. Let $P_{\mathsf{max}}$ denote the maximum transmit power of each AP. We define the RSRP received by $\mathsf{UE}_j$ from $\mathsf{AP}_i$ as $\mathsf{RSRP}_{ji}=P_{\mathsf{max}}H_{ji}^2$, which approximates the average power that $\mathsf{UE}_j$ receives from $\mathsf{AP}_i$ over time. We then define the set of UEs associated with $\mathsf{AP}_i$, denoted by $\mathcal{L}_i$, as
\begin{align}\label{eq:def_UEpool}
\mathcal{L}_i = \left\{ \mathsf{UE}_j: i = \arg \max_{i\in\{1,\dots,N\}} \mathsf{RSRP}_{ji} \right\},
\end{align}
where ties are broken arbitrarily. Note that this association policy is equivalent to the \emph{max-SINR} (signal-to-interference-plus-noise ratio) user association policy, because if a UE receives the highest RSRP from an AP, the UE will also experience the highest average SINR from that same AP in the presence of interference from other APs~\cite{dhillon2013load}. Moreover, note that this association policy may not be optimal in general, and as mentioned in Section~\ref{sec:sys_model}, any arbitrary user association criterion may be used with our framework provided that it satisfies~\eqref{eq:user_pool1}-\eqref{eq:user_pool3}.  

\subsection{Baseline Algorithms}\label{sec:baselines}
We compare the performance of our proposed scheduler against several baseline algorithms.
\begin{itemize}[leftmargin=*]
\item \textbf{Full reuse:} At each scheduling interval, each AP schedules the UE in its local user pool with the highest PF ratio (PF-based user scheduling), and serves it with full transmit power. 

\item \textbf{Time division multiplexing (TDM):} The UEs are scheduled in a round-robin fashion. In particular, at scheduling interval $t$, $\mathsf{UE}_{t \: \mathrm{mod} \: K}$ is scheduled to be served with full transmit power by its associated AP, while the rest of the APs remain silent. 

\item \textbf{Information-theoretic link scheduling (ITLinQ)~\cite{naderializadeh2014itlinq, naderializadeh2017ultra}:} This is a centralized binary power control algorithm, in which UEs are first selected using PF-based scheduling, and then the AP-UE pairs are sorted in the descending order of the selected UEs' PF ratios. The AP whose selected UE has the highest PF ratio is scheduled to transmit, and going down the ordered list, each $\mathsf{AP}_i$ is also scheduled to serve its selected $\mathsf{UE}_{j_i}$ if for every $k$ such that $\mathsf{AP}_k$ is active,
\begin{align}\label{eq:ITLQ}
\max\left\{\frac{P_{\mathsf{max}}|h_{j_k i}|^2}{\sigma^2},\frac{P_{\mathsf{max}}|h_{j_i k}|^2}{\sigma^2}\right\} < M \cdot \left(\frac{P_{\mathsf{max}}|h_{j_i i}|^2}{\sigma^2} \right)^\eta,
\end{align}
where $M$ and $\eta$ are design parameters. Otherwise, $\mathsf{AP}_i$ will remain silent for that scheduling interval. The condition in~\eqref{eq:ITLQ} is inspired by the information-theoretic condition for the approximate optimality of treating interference as noise~\cite{geng2015optimality}, ensuring that the interference-to-noise ratios (INRs), both caused by $\mathsf{AP}_i$ at already-scheduled UEs and received by $\mathsf{UE}_{j_i}$ from already-scheduled APs, are ``weak enough'' compared to the signal-to-noise ratio (SNR) between $\mathsf{AP}_i$ and $\mathsf{UE}_{j_i}$. For our simulations, we consider $M=1$ and $\eta=0.4$. 

\end{itemize}

\begin{figure*}[t]
\centering
\setlength{\belowcaptionskip}{-10pt}
\includegraphics[width=.31\textwidth]{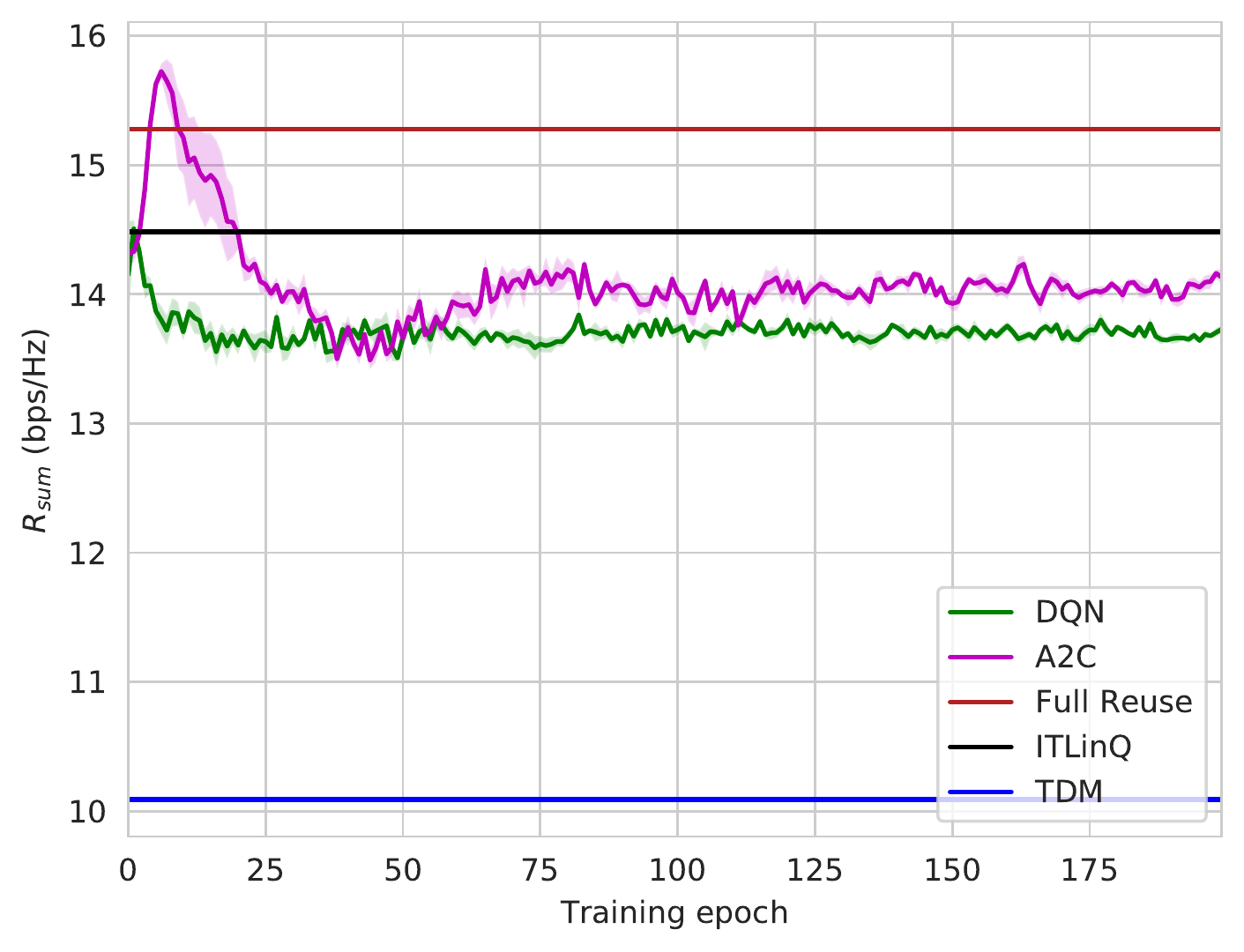}
~
\includegraphics[width=.32\textwidth]{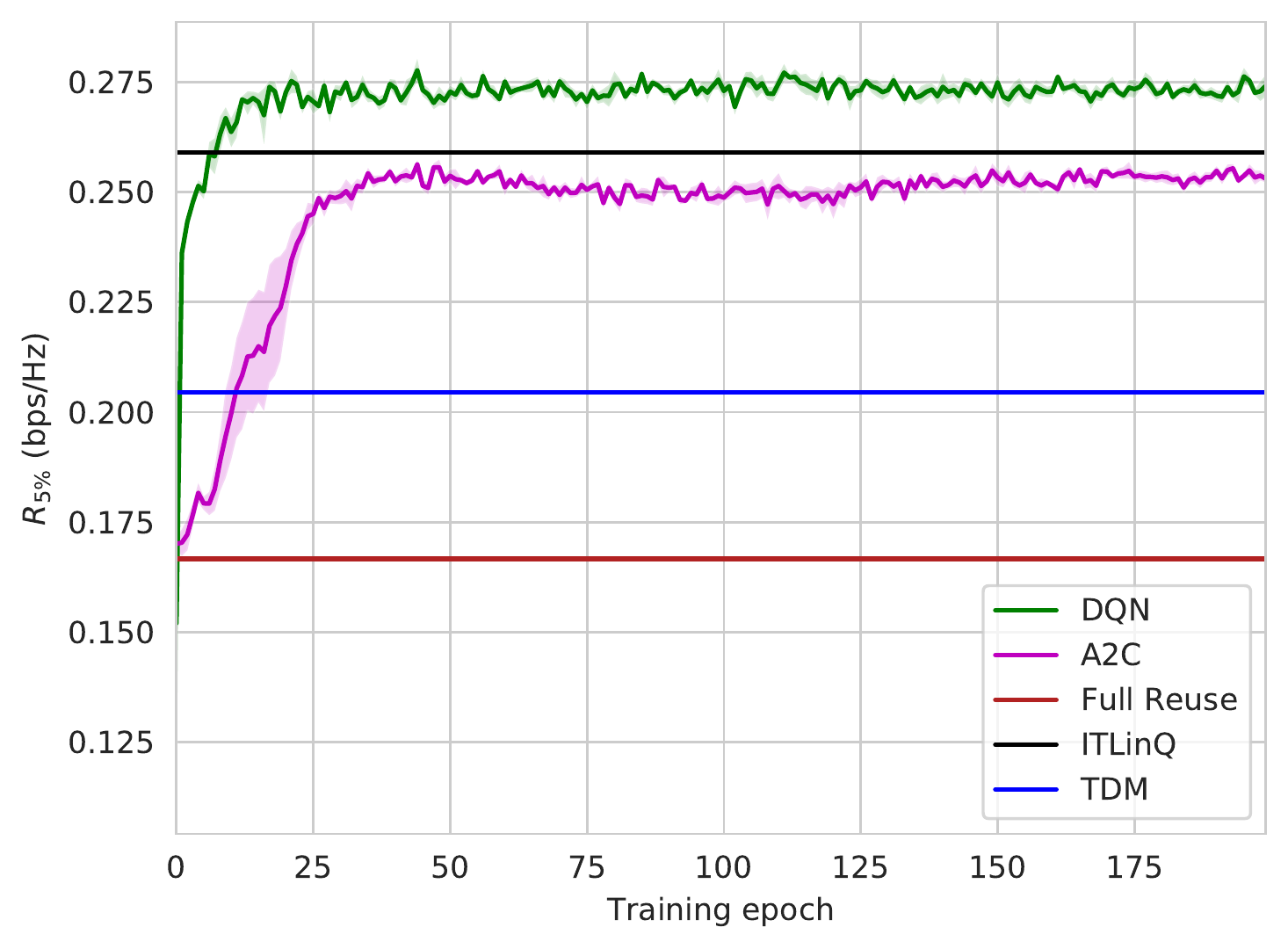}
~
\includegraphics[width=.31\textwidth]{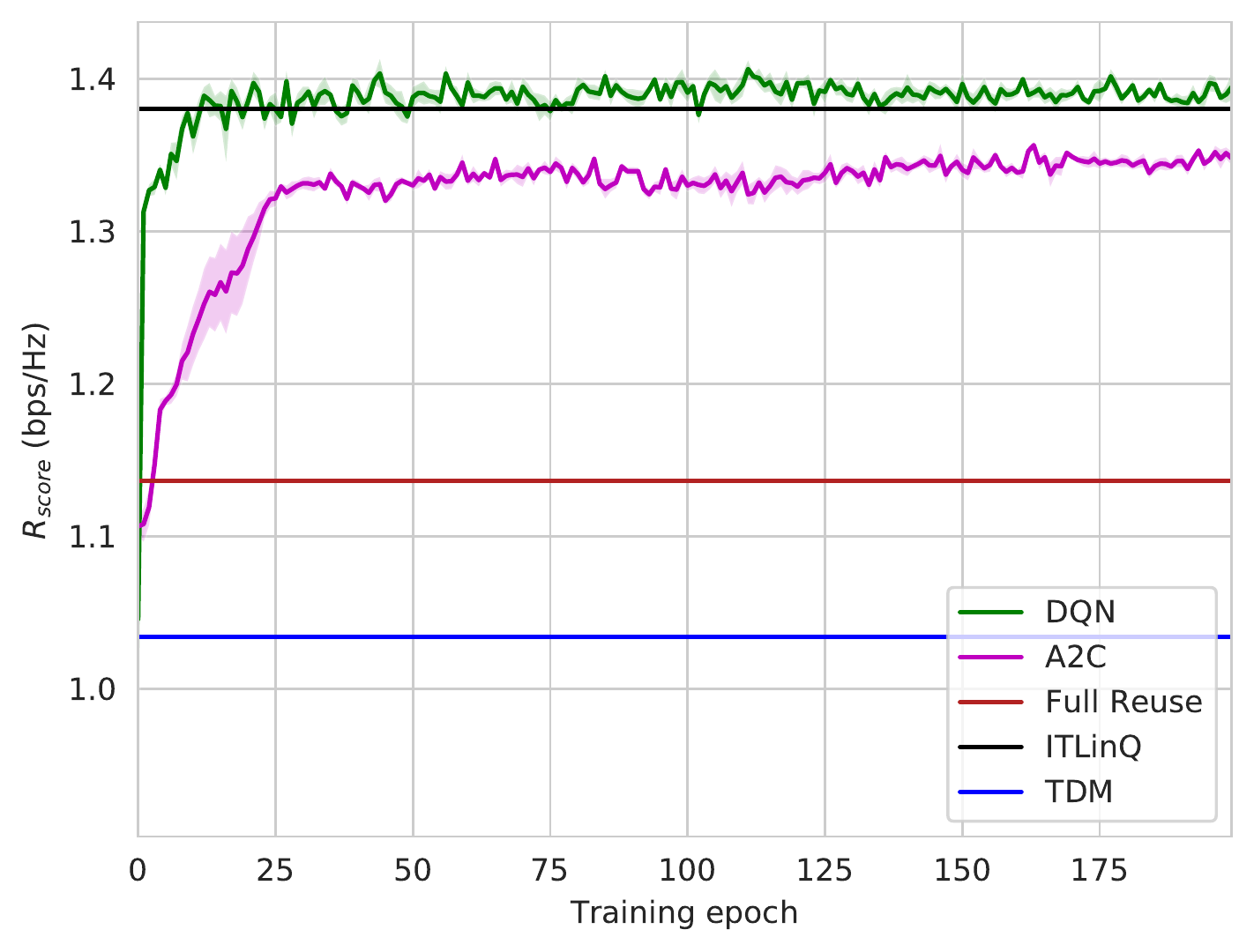}
\caption{The sum-rates (left), 5\textsuperscript{th} percentile rates (middle), and scores (right) achieved by the DQN and A2C agents alongside baseline algorithms over the 50 validation environments, when training on environments with 4 APs and 24 UEs.}
\label{fig:training_4x24}
\end{figure*}

\subsection{Deep MARL Agents}

We consider the following two different types of agents with the (hyper)parameters in Table~\ref{table:agent_params}:
\begin{itemize}[leftmargin=*]
\item \textbf{Double Deep-Q Network (DQN)~\cite{mnih2015human,van2016deep}:} We consider a double DQN agent, which is a value-based model-free deep RL method. We use the concurrent sampling technique proposed in~\cite{omidshafiei2017deep}, where the experiences of all agents at a given scheduling interval are sampled together from the experience buffer, allowing us to capture the correlation between the agents' concurrent experiences and the impact of their joint actions on their rewards.

\item \textbf{Advantage Actor-Critic (A2C)~\cite{wang2016learning}:} We use the OpenAI baselines~\cite{baselines} implementation of an A2C agent---which is a policy-based model-free deep RL method and a synchronous version of the asynchronous advantage actor-critic (A3C) agent~\cite{mnih2016asynchronous}.
\end{itemize}

To improve the agents' generalization capabilities, we use $L_2$ regularization on the agents' parameters~\cite{cobbe2018quantifying}. We train each of the agents for $2000$ episodes, or $200$ epochs, where an epoch is defined as a group of $10$ consecutive episodes. To evaluate our agents during training, we create a set of $50$ \emph{validation environments}, whose average and 5\textsuperscript{th} percentile rates are within a $5\%$ relative error of those achieved over 1000 random environment realizations by both full reuse and TDM baseline algorithms. At the end of each training epoch, the trained agents are tested across the validation environments. Once training is complete, the resulting models are evaluated on another randomly-generated set of 1000 environment realizations.

For each type of environment configuration, we train 5 models, utilizing different random number generator seeds. In the following sections, we report the mean of the results across the trained 5 models, with the shaded regions around the curves illustrating the standard deviation.

\subsection{Validation Performance during Training}\label{sec:train_results}

We first demonstrate how the behavior of the model evolves as training proceeds. Figure~\ref{fig:training_4x24} illustrates the evolution of validation sum-rate, 5\textsuperscript{th} percentile rate, and a score metric $R_{\mathsf{score}}$, defined as a score metric $R_{\mathsf{score}}$, defined as\footnote{We have used the factor of 3 for the 5\textsuperscript{th} percentile rate in defining $R_{\mathsf{score}}$, because prior experience has shown that improving cell-edge performance is typically three times more challenging than enhancing cell-center performance.} $R_{\mathsf{score}} \triangleq \frac{R_{\mathsf{sum}}}{K} + 3 \times R_{5\%}$, when training on environments with $N=4$ APs and $K=24$ UEs.

As the plots in Figure~\ref{fig:training_4x24} show, both DQN and A2C agents initially favor sum-rate performance, while suffering in terms of the 5\textsuperscript{th} percentile rate. As training proceeds, the agents learn a better balance between the two metrics, trading off sum-rate for improvements in terms of 5\textsuperscript{th} percentile rate. As the figure shows, A2C achieves a better sum-rate, while DQN achieves a better coverage and also a better score, outperforming the centralized ITLinQ approach after only 12 epochs. Moreover, DQN converges faster than A2C, due to better sample efficiency thanks to the experience buffer. Note that the centralized ITLinQ approach conducts link scheduling based on the condition for the \emph{approximate} optimality of treating interference as noise, and therefore does not guarantee optimality in general.

For each training run, we select the model with the highest score level, and then use it to conduct final, large-scale, test evaluations upon the completion of training, as we will show in the next section.

\begin{remk}
Figure~\ref{fig:SINR_vs_num_remotes} shows the variation in the mean long-term UE SINR in networks with $4-10$ APs and 100 UEs, where for each configuration of $N$ APs, the interference includes the contribution from $n\in\{0, \dots, N-1\}$ remote agents physically-closest to the serving AP.

\begin{figure}[t]
\centering
\setlength{\belowcaptionskip}{-10pt}
\includegraphics[width=.8\linewidth]{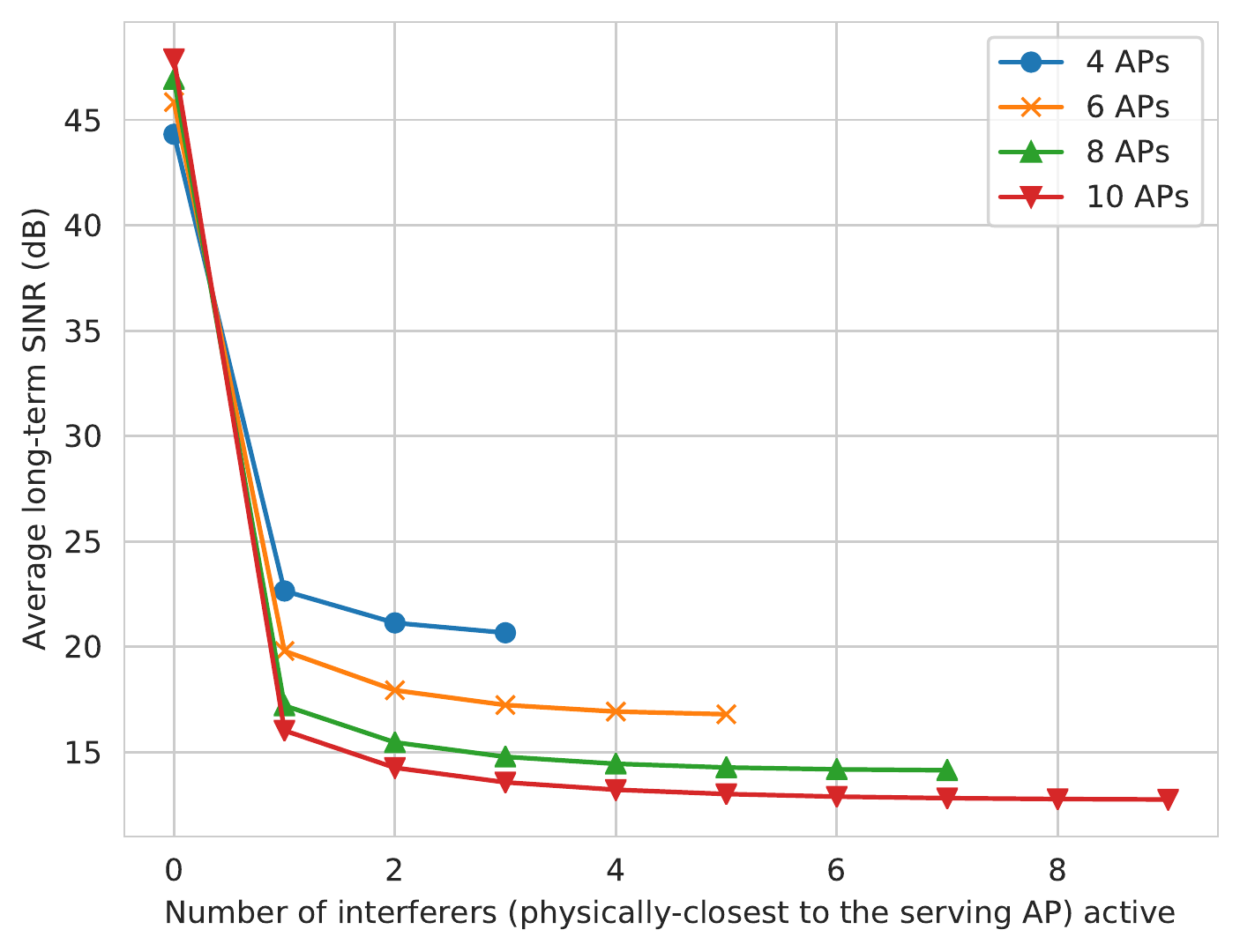}%
\caption{Impact of the number of strongest interferers included in the interference calculation on the average long-term SINR of UEs in networks with $N\in\{4,6,8,10\}$ APs and $100$ UEs.}
\label{fig:SINR_vs_num_remotes}
\end{figure}

This figure sheds light on how many remote agents' observations should be included in each agent's observations. As the figure demonstrates, by far the largest reduction in SINR occurs when the closest AP transmits. Moreover, the curves flatten out as the number of included interference terms increases, indicating that interference from farther APs is less consequential and may be safely omitted from the observation space.
\end{remk}

\begin{figure*}[t]
\centering
\setlength{\belowcaptionskip}{-5pt}
\includegraphics[width=.45\textwidth]{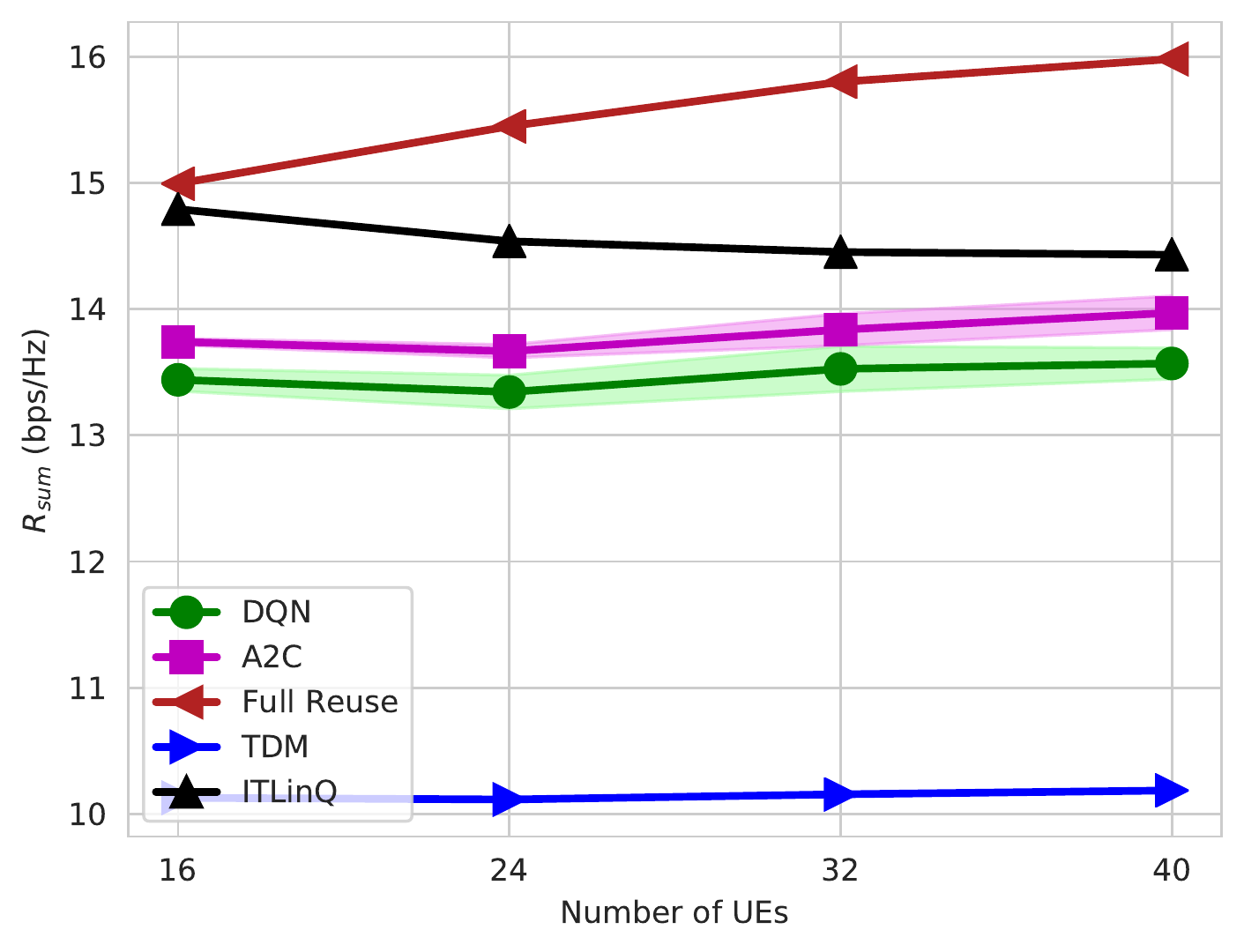}
\qquad
\includegraphics[width=.465\textwidth]{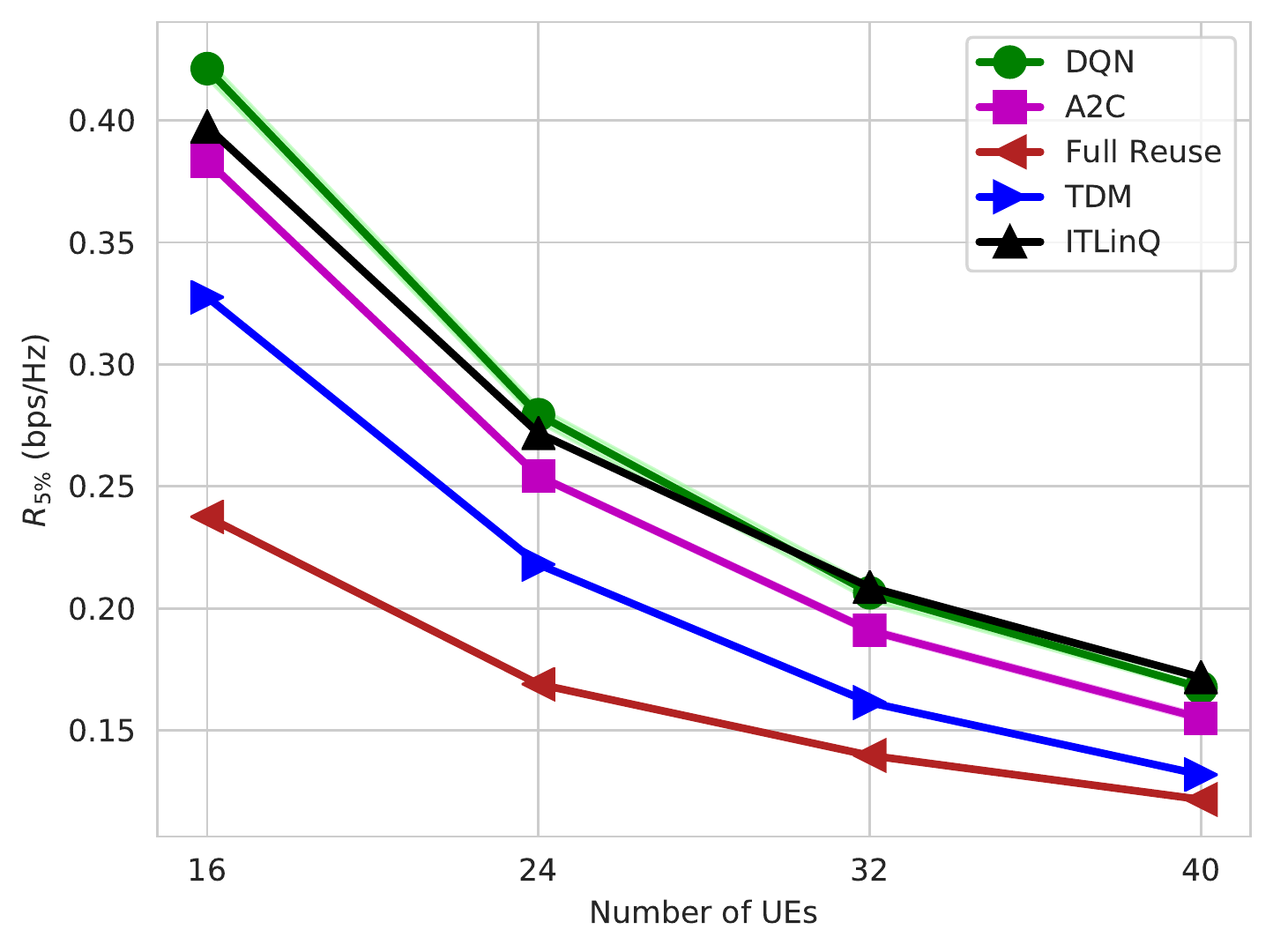}
\caption{Comparison of the achievable sum-rates (left) and 5\textsuperscript{th} percentile rates (right) of models trained on environments with 4 APs and various numbers of UEs with those of baseline algorithms, where the model trained on each configuration was deployed on the same configuration during the test phase.}
\label{fig:self_4TP}
\end{figure*}

\begin{figure*}[t]
\centering
\setlength{\belowcaptionskip}{-5pt}
\includegraphics[width=.45\textwidth]{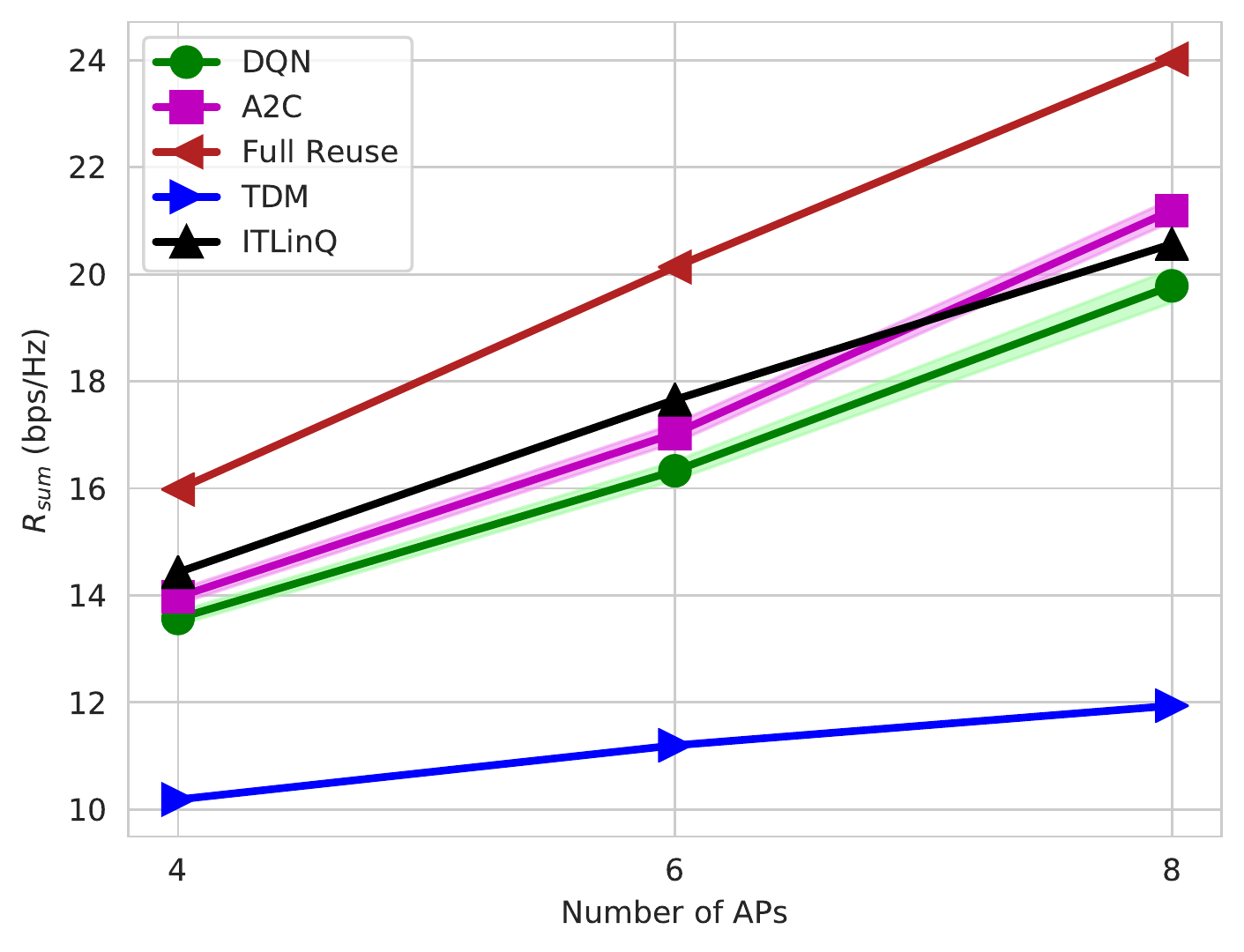}
\qquad
\includegraphics[width=.465\textwidth]{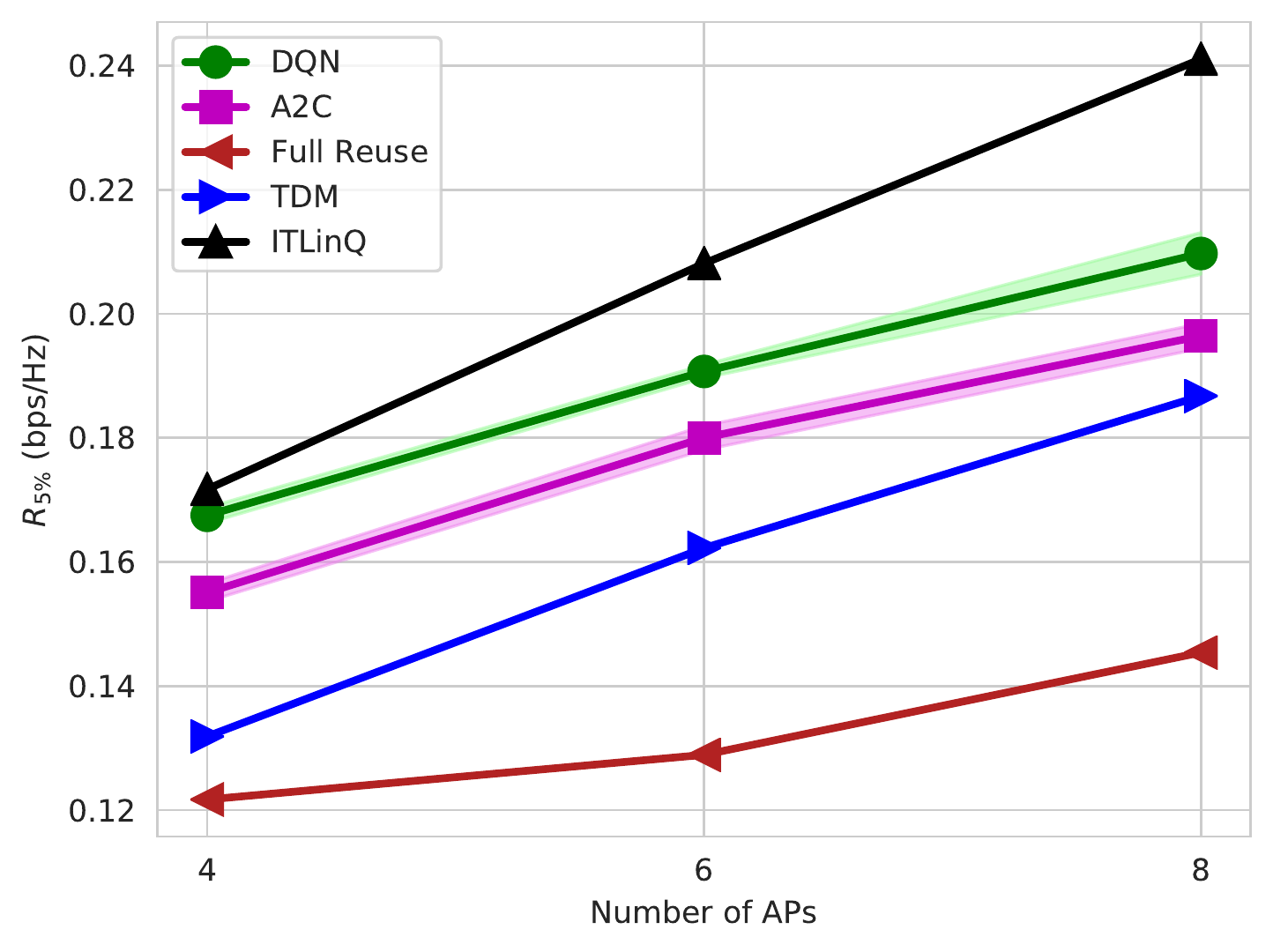}
\caption{Comparison of the achievable sum-rates (left) and 5\textsuperscript{th} percentile rates (right) of models trained on environments with 40 UEs and various numbers of APs with those of baseline algorithms, where the model trained on each configuration was deployed on the same configuration during the test phase.}
\label{fig:self_xTP}
\end{figure*}

\subsection{Final Test Performance with Similar Train and Test Configurations}\label{sec:self_tests}

In this section, we present the final test results for models tested on the same configuration as the one in their training environment. Figure~\ref{fig:self_4TP} demonstrates the achievable sum-rate and 5\textsuperscript{th} percentile rate for the environments with 4 APs and varying numbers of UEs. As the plots show, our proposed deep RL methods significantly outperform TDM in both sum-rate and 5\textsuperscript{th} percentile rate, and they also provide considerable 5\textsuperscript{th} percentile rate gains over full reuse. Our reward design helps the agents achieve a balance between sum-rate and 5\textsuperscript{th} percentile rate, helping the DQN agent attain 5\textsuperscript{th} percentile rate values which are on par with ITLinQ for large numbers of UEs (32-40), while outperforming it for smaller numbers of UEs (16-24). The A2C agent, on other hand, performs consistently well in terms of the sum-rate, approaching ITLinQ as the number of users increases across the network.

In Figure~\ref{fig:self_xTP}, we plot the sum-rate and 5\textsuperscript{th} percentile rate for the configurations with 40 UEs and different numbers of APs. As the figure shows, the relative trends are similar to the previous case in terms of sum-rate, with A2C outperforming ITLinQ for networks with 8 APs, but in terms of the 5\textsuperscript{th} percentile rate, both agents outperform TDM and full reuse, while having inferior performance relative to the centralized ITLinQ approach as the number of APs, and equivalently the number of agents, gets larger.

\begin{figure*}[t]
\centering
\setlength{\belowcaptionskip}{-8pt}
\includegraphics[width=.45\textwidth]{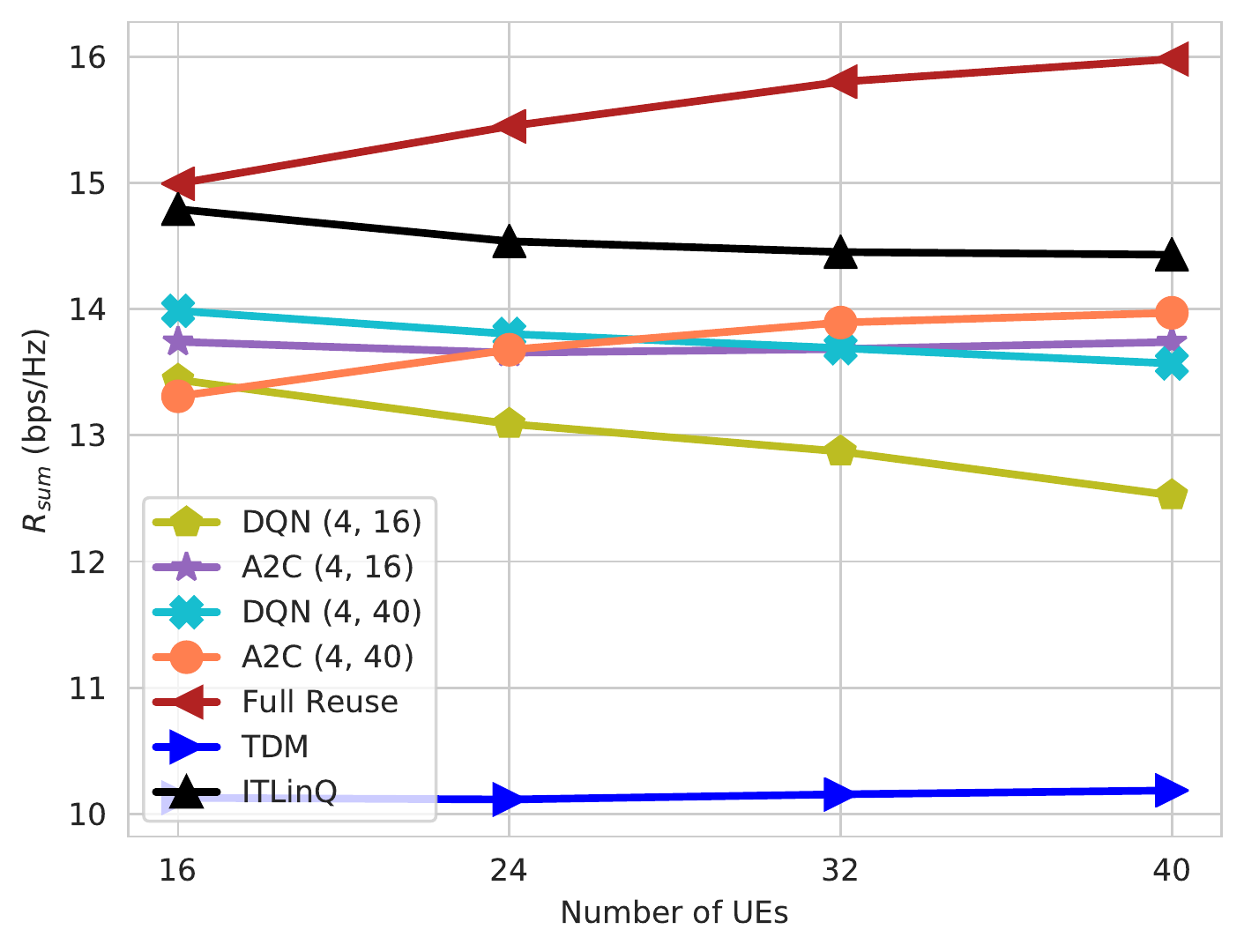}
\qquad
\includegraphics[width=.465\textwidth]{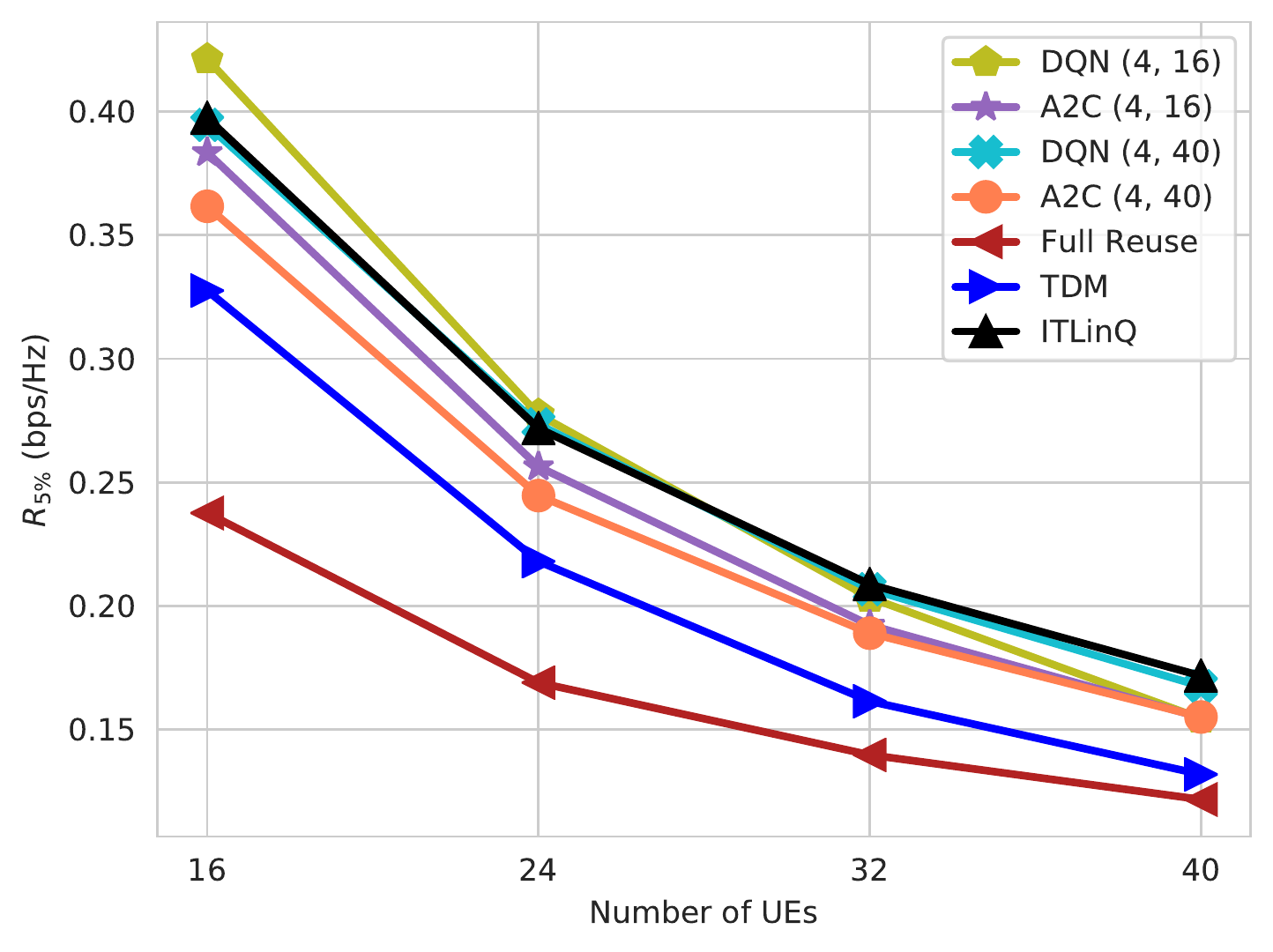}
\caption{Comparison of the achievable sum-rates (left) and 5\textsuperscript{th} percentile rates (right) of models trained on environments with 4 APs and various numbers of UEs with those of baseline algorithms, where the model trained on each configuration was deployed on all the other configurations as well. The first and second element of each tuple in the legends represent numbers of APs and UEs in the training environment, respectively.}
\label{fig:cross_4TP}
\end{figure*}

\begin{figure*}[t]
\centering
\setlength{\belowcaptionskip}{-5pt}
\includegraphics[width=.45\textwidth]{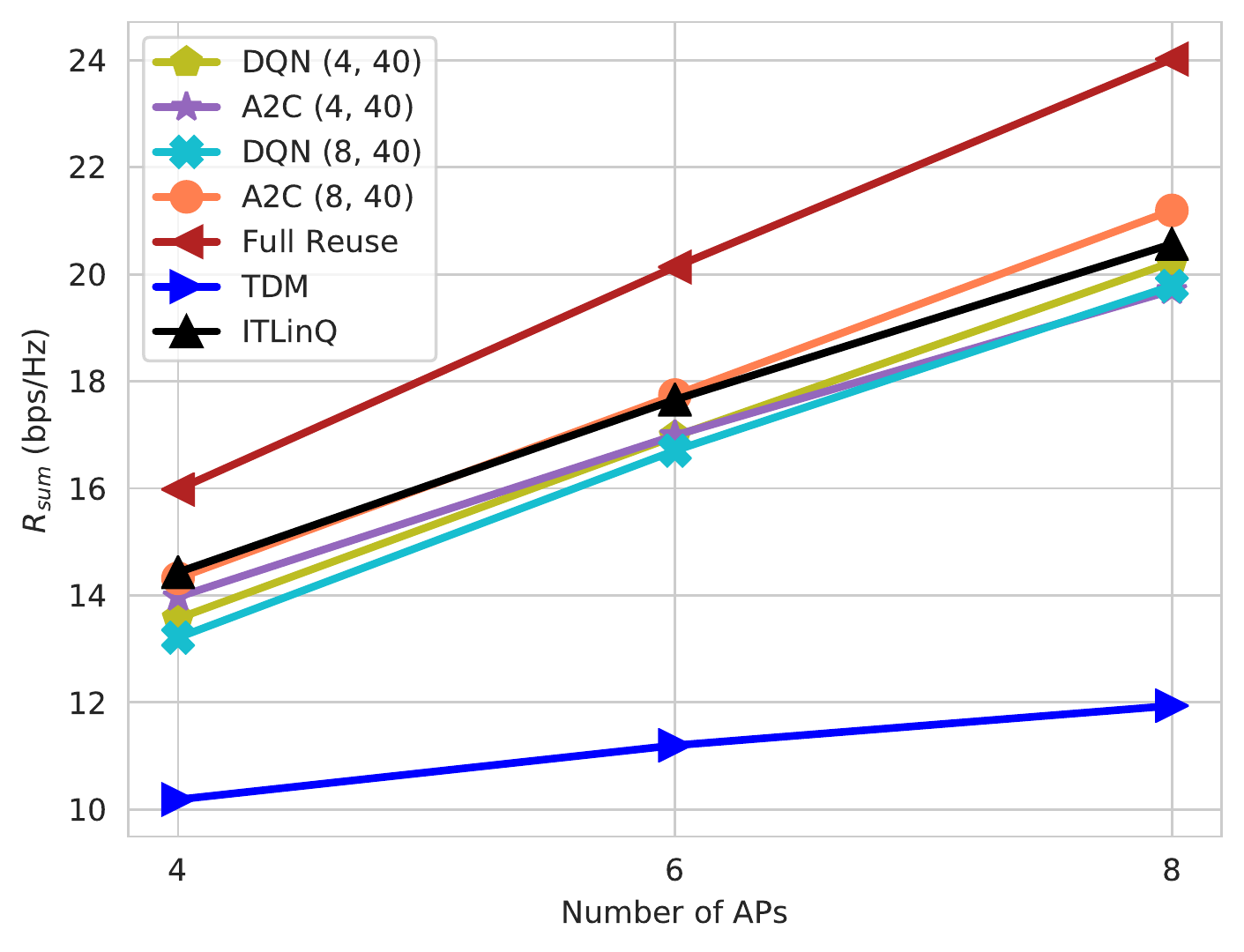}
\qquad
\includegraphics[width=.465\textwidth]{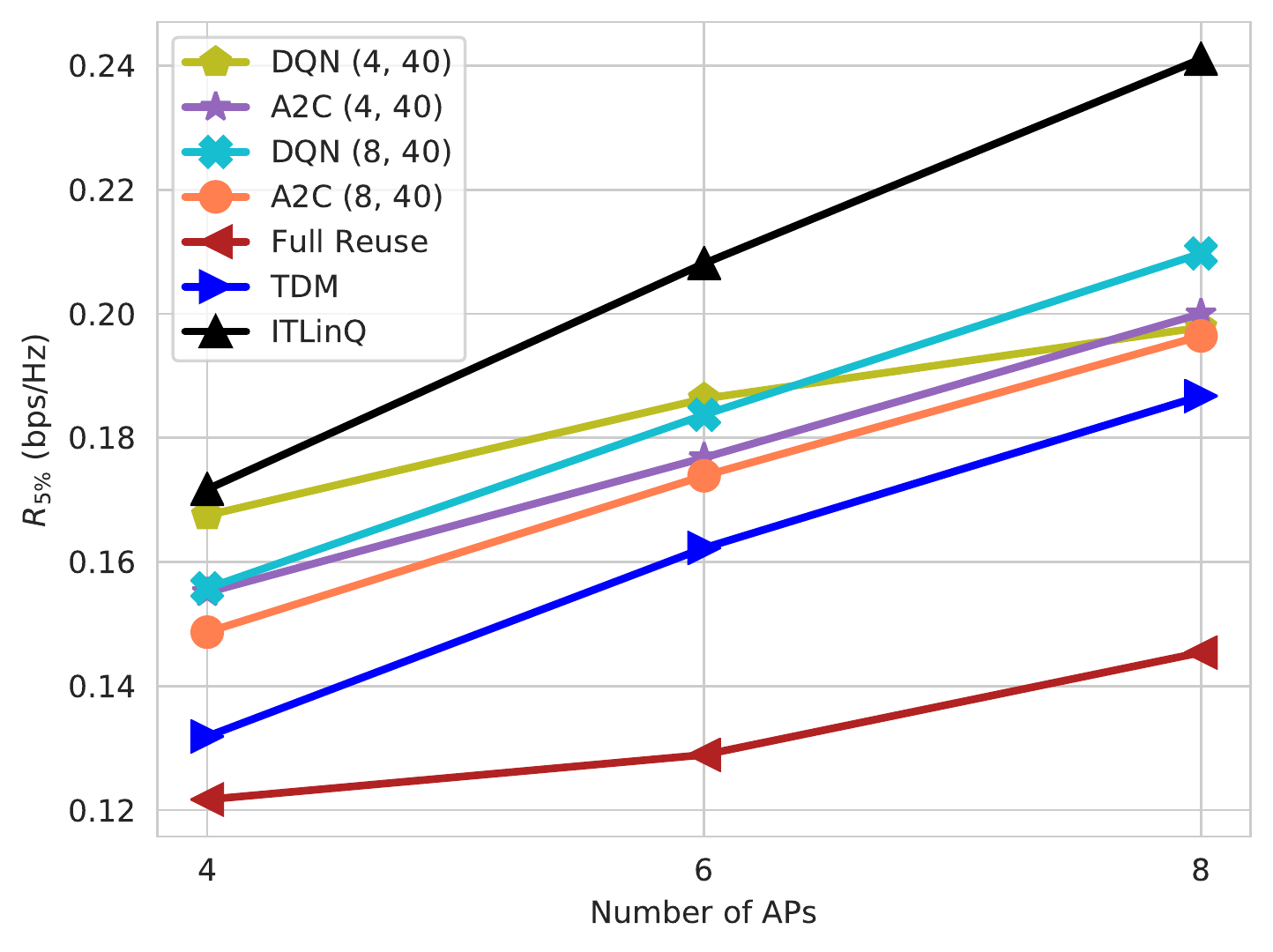}
\caption{Comparison of the achievable sum-rates (left) and 5\textsuperscript{th} percentile rates (right) of models trained on environments with 40 UEs and various numbers of APs with those of baseline algorithms, where the model trained on each configuration was deployed on all the other configurations as well. The first and second element of each tuple in the legends represent numbers of APs and UEs in the training environment, respectively.}
\label{fig:cross_xTP}
\end{figure*}

\subsection{Final Test Performance with Discrepant Train and Test Configurations}\label{sec:cross_tests}
As mentioned before, our design of the observation and action spaces is such that they have a fixed size regardless of the actual training configuration. We test the robustness of our models with respect to network density by testing policies trained on one density deployed in environments of other densities. To reduce clutter, in the following, we only plot the average results over the 5 seeds and remove the shaded regions representing the standard deviation of the results.

We first test models trained on environments with $N=4$ APs and different numbers of UEs against each other, and plot the results in Figure~\ref{fig:cross_4TP}. We observe that all DQN and A2C agents are robust in terms of both metrics. Interestingly, the A2C model trained on the case with 16 UEs has a much better performance (especially in terms of sum-rate) than its counterpart DQN model as the number of UEs in the test deployment increases. For models with 40 UEs, however, DQN tends to perform better, especially in terms of the 5\textsuperscript{th} percentile rate.

Next, we cross-test the models trained on environments with 40 UEs and different numbers of APs against each other. Figure \ref{fig:cross_xTP} shows the sum-rates and 5\textsuperscript{th} percentile rates achieved by these models. All the models exhibit fairly robust behaviors with the exception of the DQN model trained on configurations with 4 APs, whose 5\textsuperscript{th} percentile rate performance deteriorates for higher numbers of APs. Note that in this case, the number of agents changes across different scenarios, and we observe that in general, training with more agents leads to more capable models, which can still perform well when deployed in sparser scenarios, while training with few agents may not scale well as the number of agents increases.

\begin{figure}[t]
\centering
\includegraphics[width=.83\linewidth]{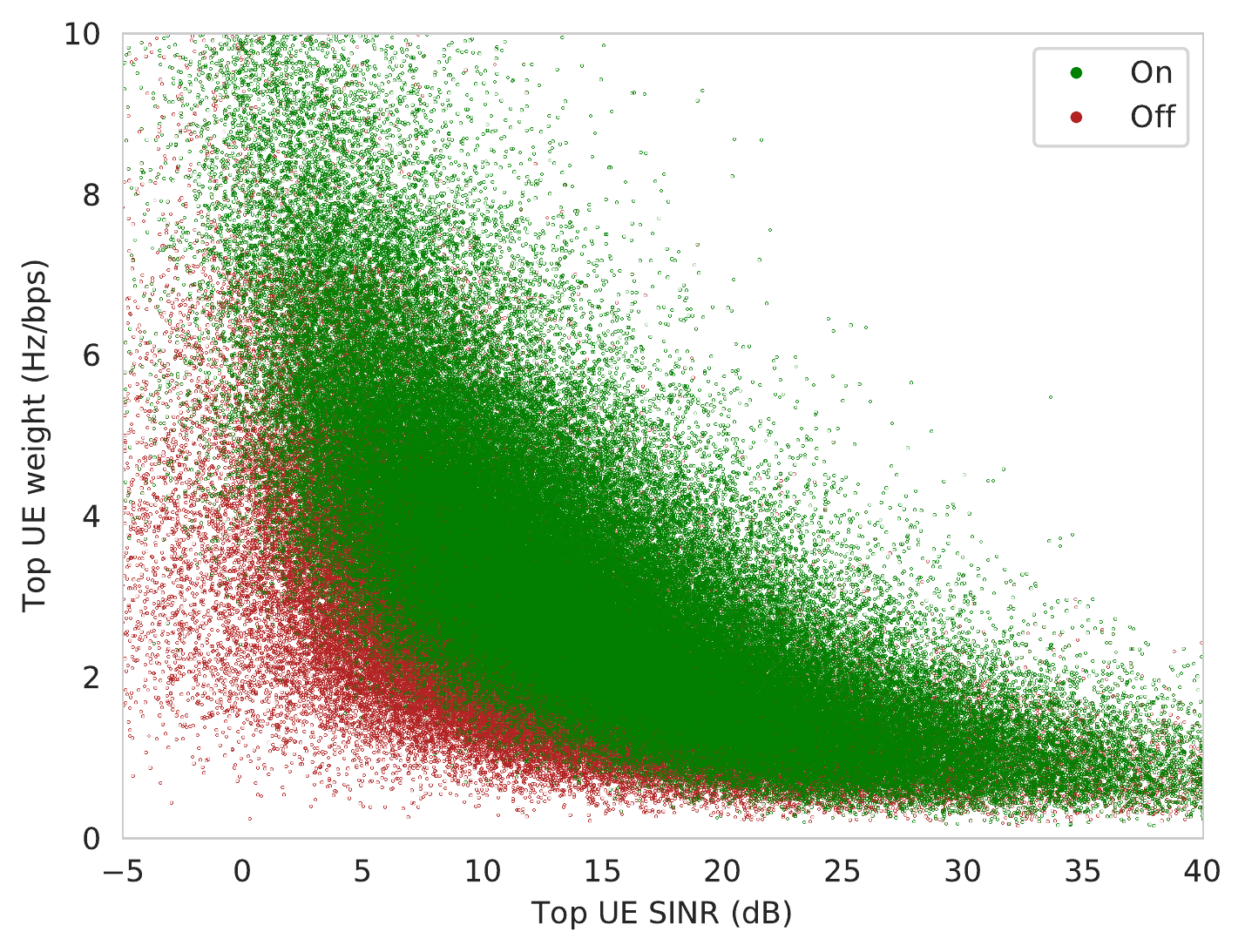}
\caption{Weight vs. SINR scatter plot of the top UE of a DQN agent trained and tested on networks with 4 APs and 24 UEs. Red (resp., green) points represent the scenarios where the agent decided to stay silent (resp., serve one of its top-3 UEs).}
\label{fig:interpret_w_vs_SINR}
\end{figure}

\begin{remk}
We have also tested our trained models with observations mapped using 20 percentile levels on test environments in which the observations were mapped using different numbers of percentile levels. For the scenario with $N=4$ APs and $K=24$ UEs, we observed that using 10-100 percentile levels during the test phase achieves results very similar to (within $3\%$ of) the ones obtained using 20 percentile levels. This shows that our proposed approach is very robust to the granularity of mapping the observations fed into the agent's neural network.
\end{remk}

\begin{figure*}[t]
\centering
\setlength{\belowcaptionskip}{-10pt}
\includegraphics[width=.315\textwidth]{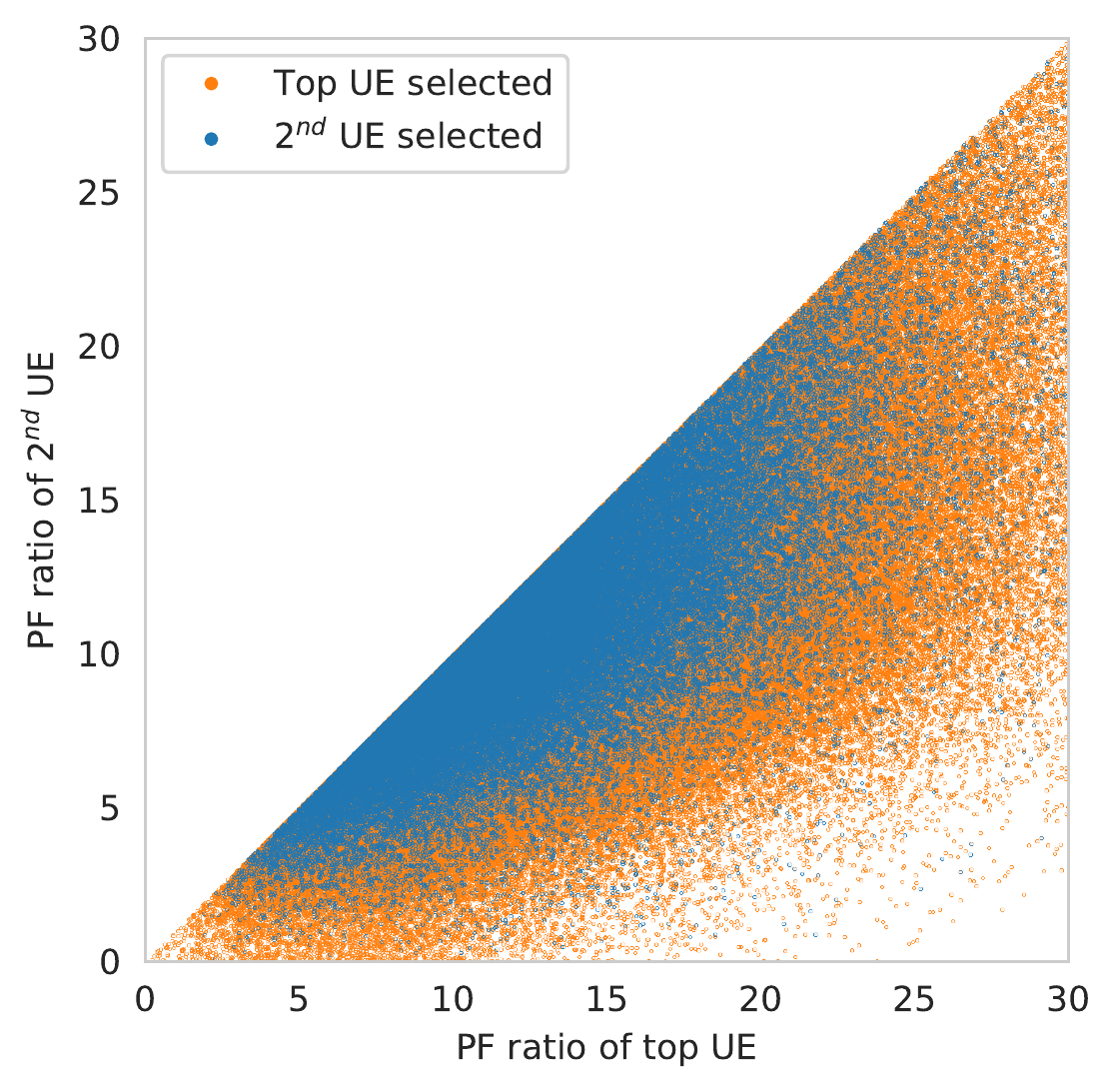}
~
\includegraphics[width=.315\textwidth]{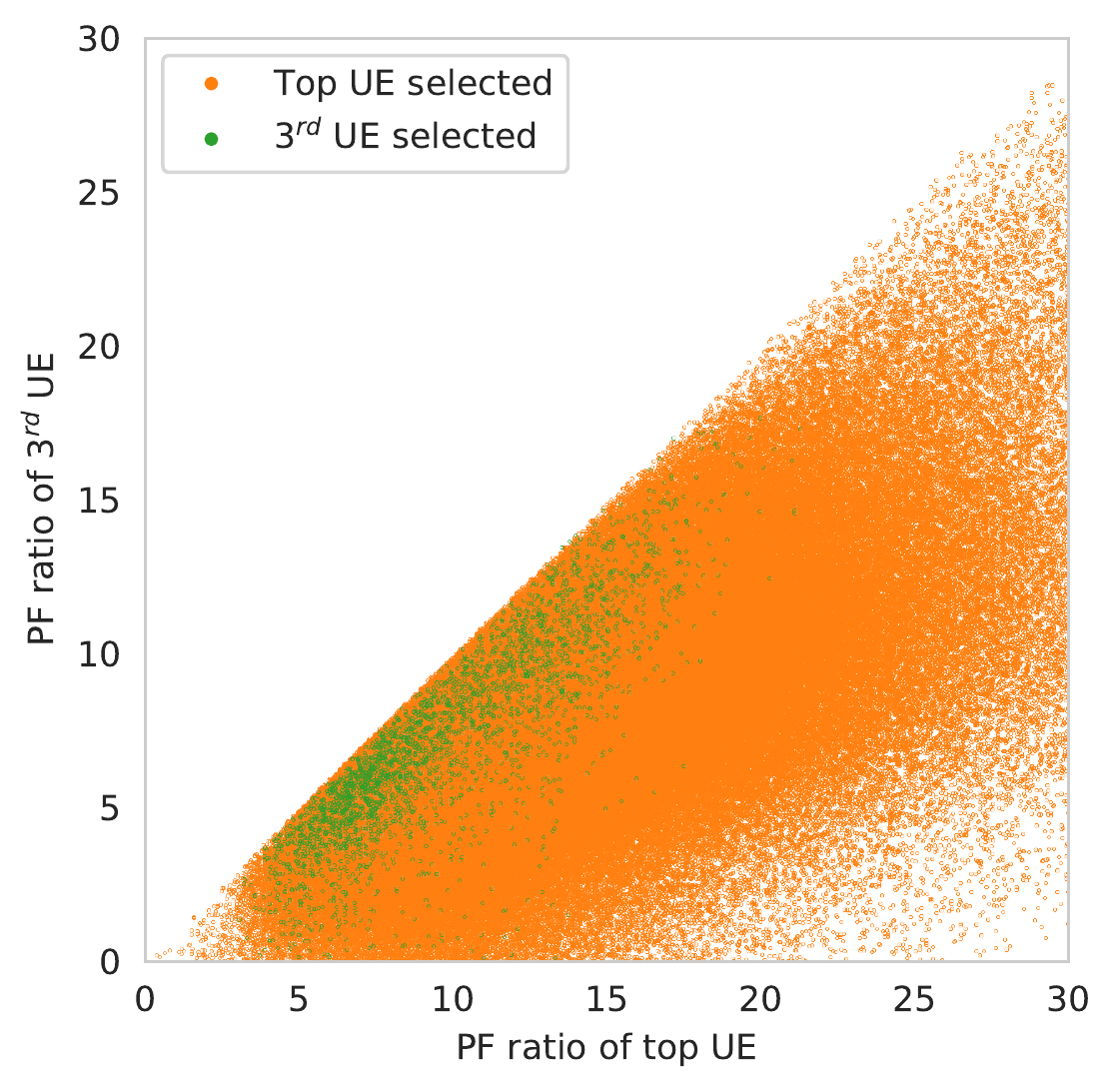}
~
\includegraphics[width=.315\textwidth]{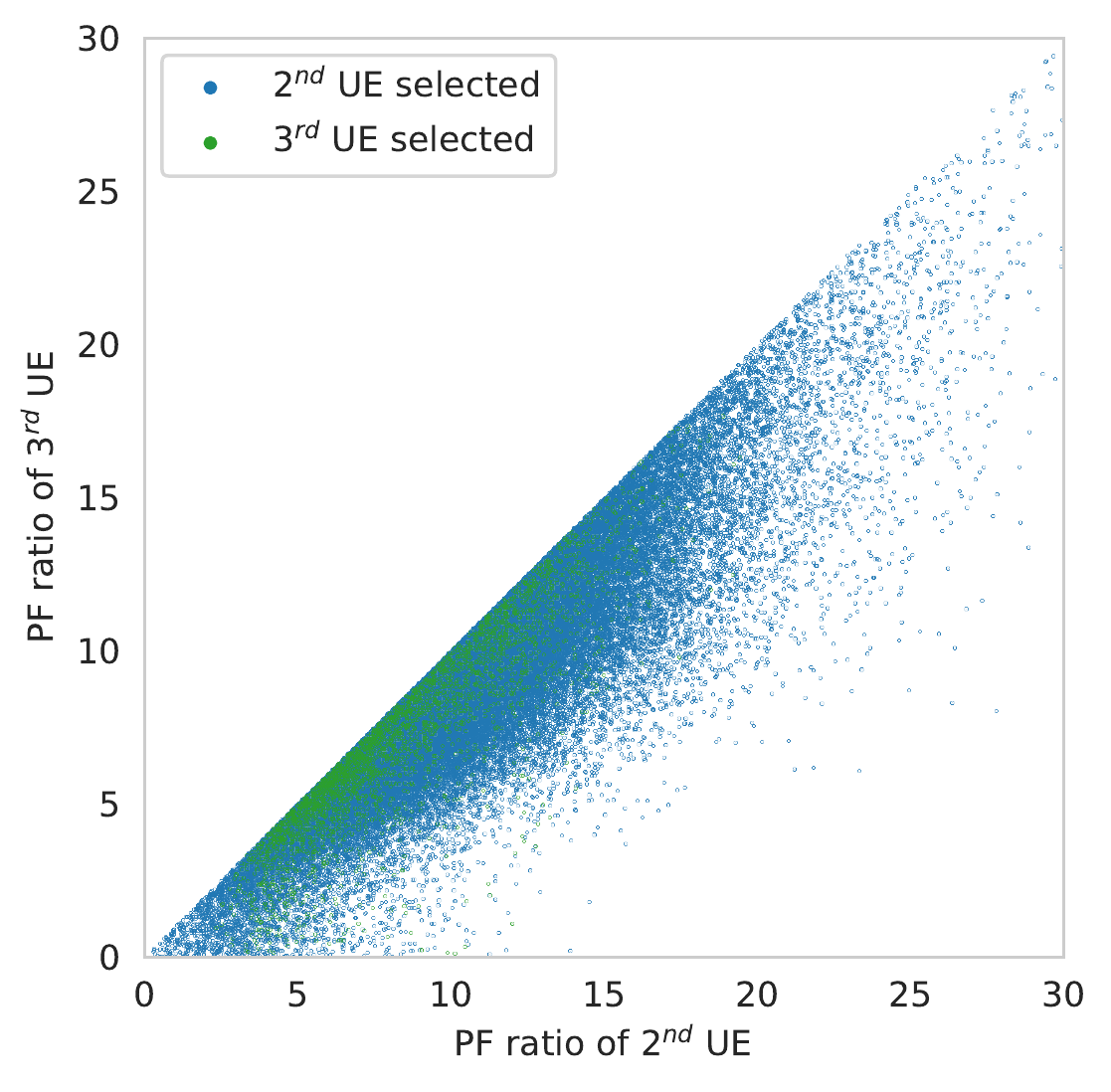}
\caption{Comparison of the PF ratios of the top-3 UEs observed by a DQN agent trained and tested on networks with 4 APs and 24 UEs for the cases in which the agent decided to serve one of those UEs.}
\label{fig:interpret_local_PFs}
\end{figure*}

\begin{figure*}[t]
\centering
\setlength{\belowcaptionskip}{-15pt}
\includegraphics[width=.315\textwidth]{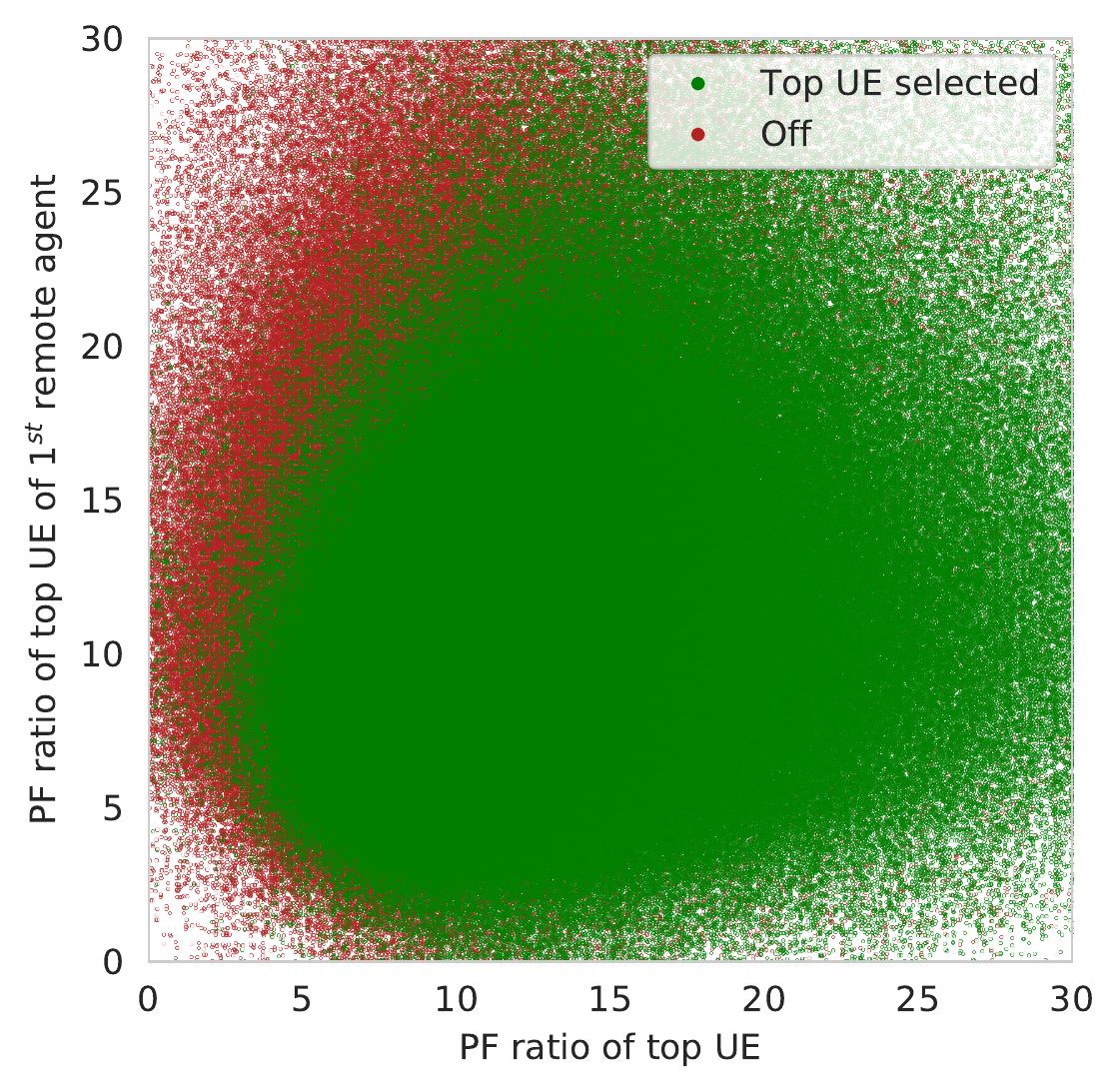}
~
\includegraphics[width=.315\textwidth]{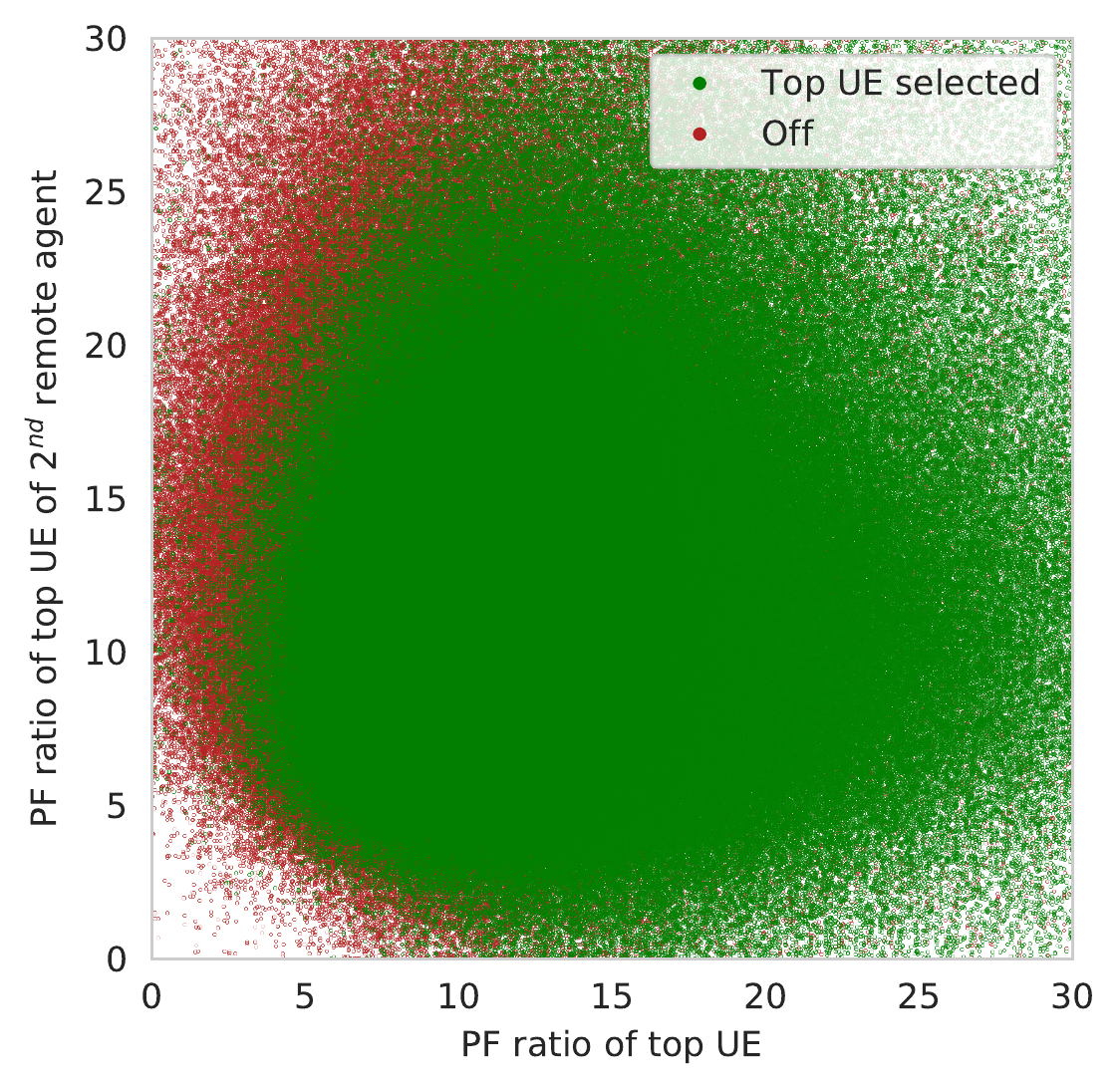}
~
\includegraphics[width=.315\textwidth]{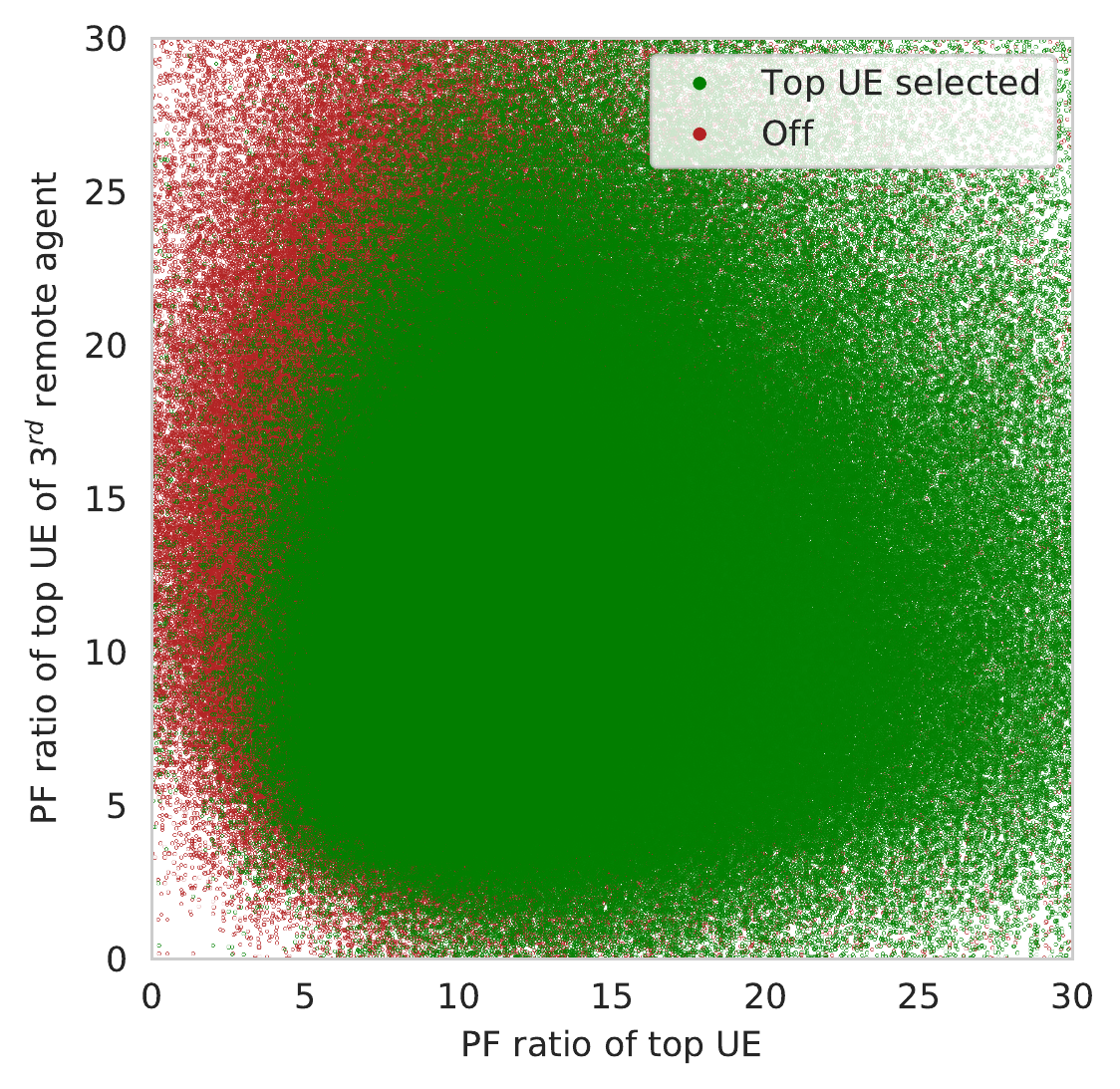}
\caption{Comparison of the PF ratios of the DQN agent's top UE and the remote agents' top UEs for an agent trained and tested on networks with 4 APs and 24 UEs. Red (resp., green) points represent the scenarios where the agent decided to remain silent (resp., serve its top UE).}
\label{fig:interpret_remotePFs}
\end{figure*}

\subsection{Interpreting Agent's Decisions}

In this section, we attempt to interpret our trained agent's decisions during the test phase. In particular, we collect data on the inputs and outputs of a DQN agent, trained on a network with $N=4$ APs and $K=24$ UEs and tested on the same configuration. Using this data, we will try to visualize the agent's actions in different situations.

Figure~\ref{fig:interpret_w_vs_SINR} shows a scatter plot of the SINR and weight of the agent's ``top UE,'' i.e., the UE in the AP's user pool with the highest PF ratio. The red points illustrates the cases where the agent decided to remain silent, while the green points represent the cases in which the agent served one of its top-3 UEs. As expected, higher weights and/or higher SINRs lead to a higher chance of the AP not being off. Quite interestingly, the boundary between the green and red regions can be approximately characterized as $w \times \mathsf{SINR}_{\text{dB}} = \mathsf{const.}$, which is effectively a linear boundary on the PF ratio; i.e., the agent decides to be active if and only if the PF ratio of its top UE is above some threshold.

Given that the PF ratio is a reasonable indicator of the status of each UE, Figure~\ref{fig:interpret_local_PFs} compares the PF ratios of the top-3 UEs included in the agent's observation and action spaces in the cases where the agent decided to serve one of the those UEs. As the figure shows, the agent's user scheduling decision heavily depends on the relative difference between the PF ratios of the top-3 UEs. In general, the second and third UEs have some chance of being scheduled if they have a PF ratio close to that of the top UE. However, this chance is significantly reduced for the third UE, as highlighted by the regions corresponding to different user scheduling actions.

Moreover, Figure~\ref{fig:interpret_remotePFs} shows the impact of remote observations on the agent's power control decisions. In particular, the figure demonstrates the cases where the agent either remains silent (red points), or decides to serve its top UE (green points). We observe that the agent learns a non-linear decision boundary between the PF ratio of its top UE and the PF ratios of the top UE of each remote agent. Notably, the green region becomes larger as we go from the left plot to the right plot. This implies that the agent ``respects'' the PF ratio of the top UE of its closest remote agent more as compared to the second and third closest remote agents, since the interference between them tends to be stronger, hence their actions impacting each other more significantly.

\begin{figure*}[t!]
    \centering
    \captionsetup[subfigure]{oneside,margin={.55cm,0cm}}
    \begin{subfigure}[t]{0.48\textwidth}
        \centering
        \includegraphics[width=.9\textwidth]{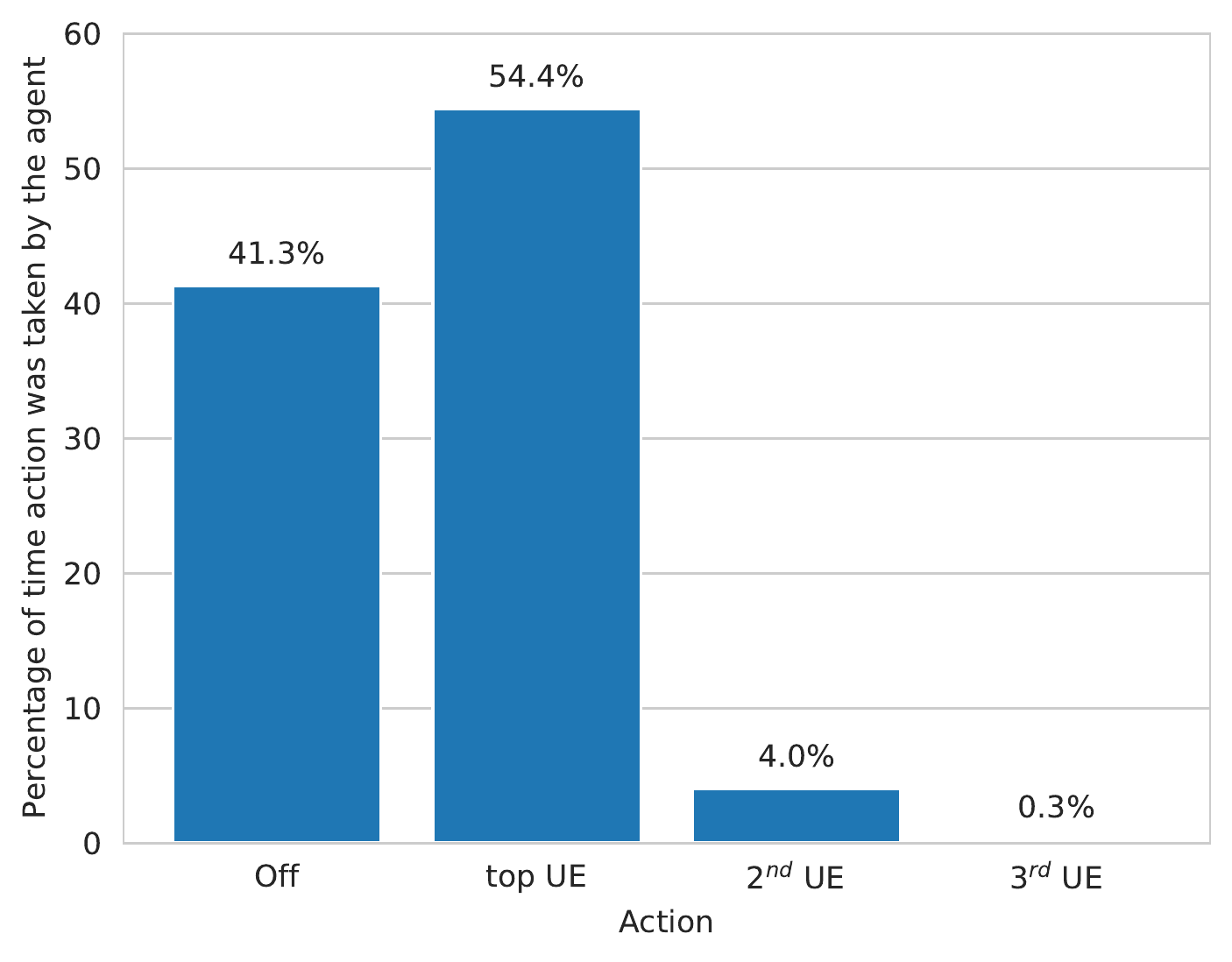}
        \caption{}
        \label{fig:action_hist_top3}
    \end{subfigure}%
    \quad 
    \begin{subfigure}[t]{0.48\textwidth}
        \centering
        \includegraphics[width=.9\textwidth]{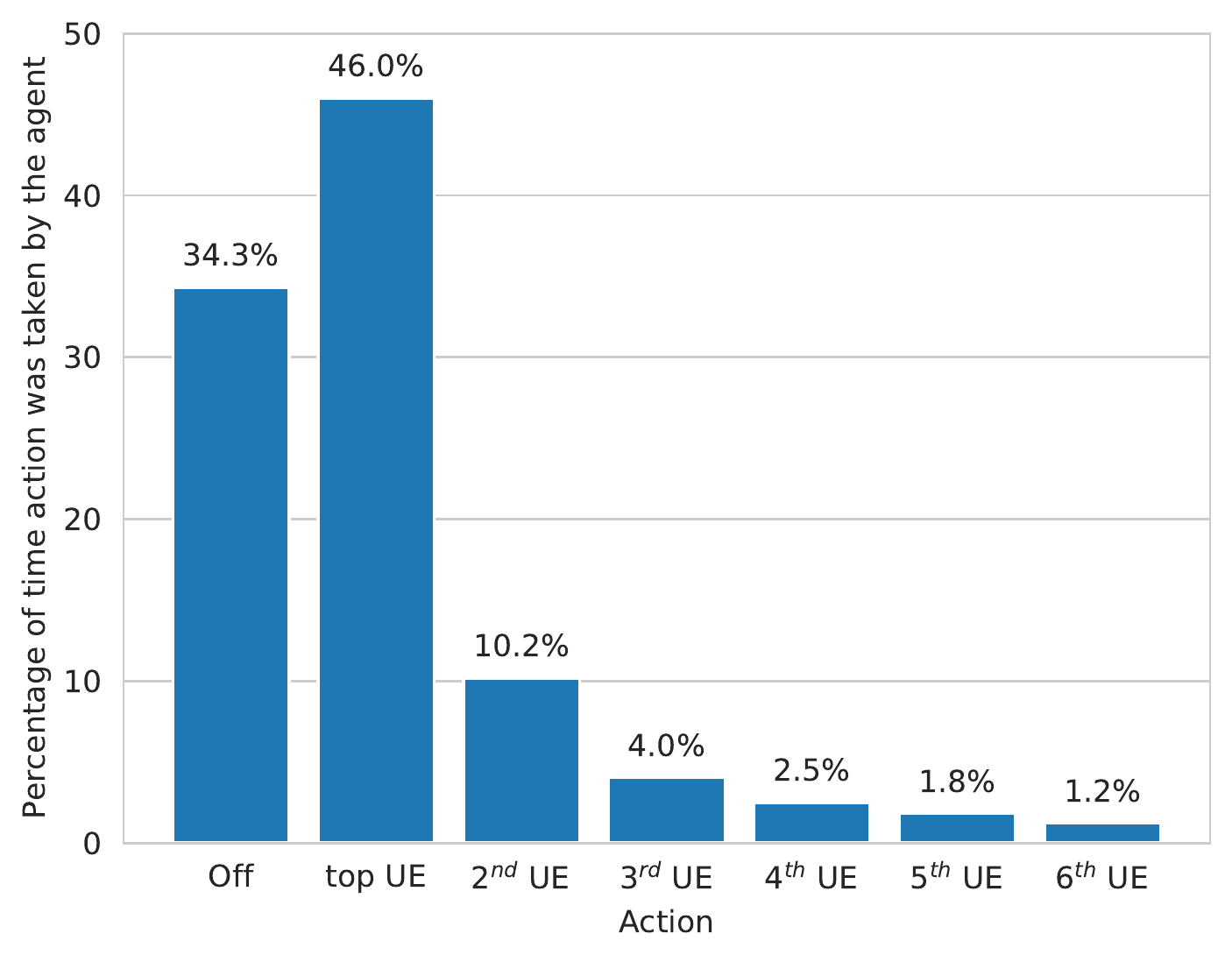}
        \caption{}
        \label{fig:action_hist_all6}
    \end{subfigure}
    \caption{Distribution of the DQN agent's actions in a network with $4$ APs and $24$ UEs, where (a) the agent observes the top-3 UEs sorted based on their PF ratios, and (b) each AP is restricted to have exactly 6 associated UEs, and the agent is able to observe all 6 UEs in an unsorted manner. For the latter case, the UE indices on the $x$-axis represent the PF-based order of the selected UE at the corresponding scheduling interval, although such an ordering was not used in the observations and actions.}
\end{figure*}

\section{Discussion}\label{sec:disc}

In this section, we discuss some of the implications of our proposed framework in more detail and provide ideas for future research on how to improve upon the current work.
\subsection{Analysis on the Number of Observable UEs by the Agent}

As described in Section~\ref{sec:RLframework}, we bound the dimension of the agent's observation and action space by selecting a finite number of $k$ UEs, whose observations are included in the agent's observation vector. We select these UEs by sorting the user pool of each AP according to their PF ratios. In this section, we shed light on the tradeoffs implied by such a ``user filtering'' method.

We first analyze the actions taken by the DQN agent when trained and tested in a network with $N=4$ APs and $K=24$ UEs. Recall that the agent can either take an ``off'' action, or decide to serve one of its top-3 UEs. We observe in Figure~\ref{fig:action_hist_top3} that the algorithm selects action 0 (no transmission) or 1 (the top UE) most of the time and rarely selects the other two UEs. In this formulation of the algorithm, including information from more than 3 UEs would most likely not improve the performance. This seems logical given that the PF ratio represents the short-term ability of the UE to achieve a high rate (represented by the SINR term) along with the long-term demand to be scheduled for the sake of fairness (represented by the weight term).

Given our goal of having a scalable agent that can be employed by any AP having an arbitrary number of associated users, it is not feasible to have an agent which can observe \emph{all} its UEs at every scheduling interval in a general network configuration. However, we conducted a controlled experiment, where we restricted the environment realizations to the ones in which all APs have a \emph{constant} number of associated UEs. In particular, we considered a configuration with $N=4$ APs and $K=24$ UEs, where each AP has exactly 6 UEs associated with it. In such a scenario, we are indeed able to design an agent, which can observe the state of \emph{all} $k=6$ of its associated UEs, as well as \emph{all} UEs associated of all its $n=3$ remote agents. Furthermore, we did \emph{not} sort the UEs of each agent according to their PF ratios. This ensures that each input port to the agent' neural network contains an observation from the same UE over time. Note that the size of the observation vector is now $2(n+1)k=48$ and the number of actions equals $1+pk=7$.

After training and testing a DQN agent on the above scenario, we observed that the resulting sum-rate and 5\textsuperscript{th} percentile rate were within 6\% of the original model using the sorted top-3 UEs. This demonstrates that the algorithm is able to learn user scheduling without the ``aid'' of sorting the UEs by PF ratio. Moreover, Figure~\ref{fig:action_hist_all6} shows the fraction of time that the model with observations from all (unsorted) UEs selects the ``off'' action and each of the UEs. For the purposes of this figure, we sorted the UEs by their PF ratios, so the percentage of time that the algorithm selects the top-UE represents the percentage of time that the agent selects the UE with the top PF ratio, regardless of its position in the observation/action space. We see that the algorithm learns to select the UE with the highest PF ratio most often, but the distribution among the various UEs is less skewed as compared to Figure~\ref{fig:action_hist_top3}. This implies that by letting the agent observe all UEs in an arbitrary order, its resulting user scheduling behavior is similar to a PF-based scheduler, but not exactly the same.

\subsection{Multiple Power Levels}\label{sec:multiple_power_levels}
\begin{figure*}[t]
\centering
\setlength{\belowcaptionskip}{-10pt}
\includegraphics[width=.45\textwidth]{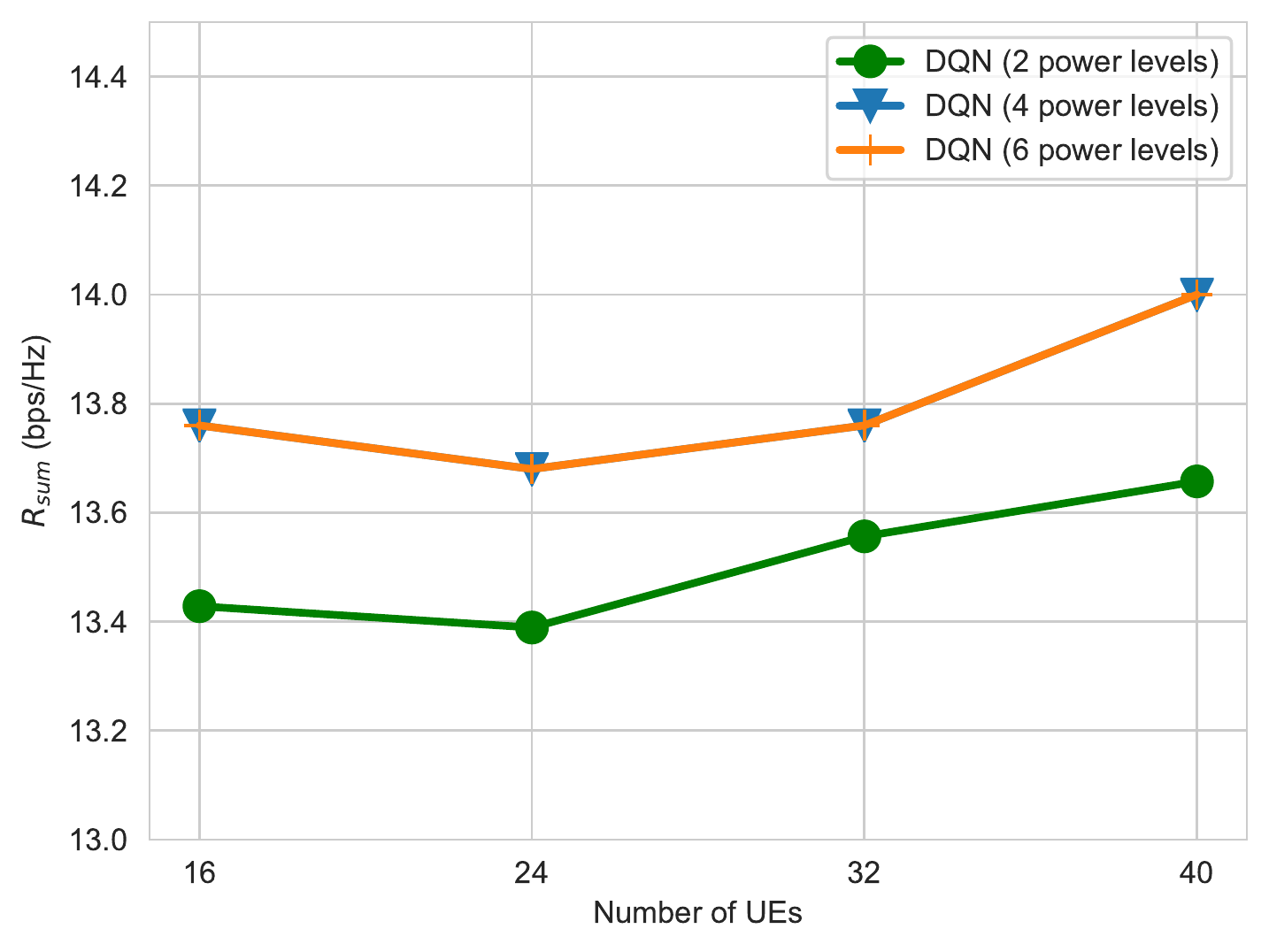}
\qquad
\includegraphics[width=.45\textwidth]{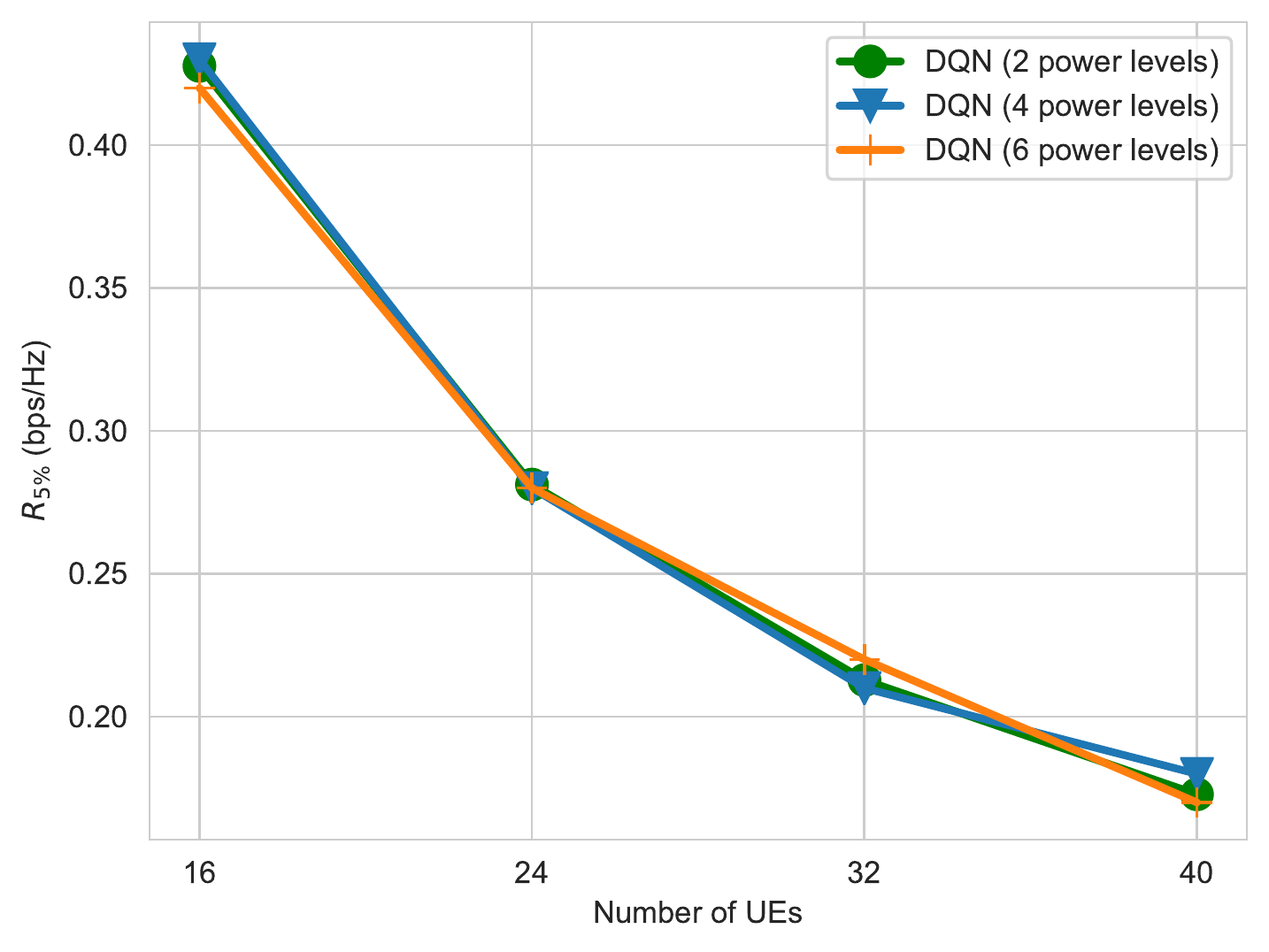}
\caption{Comparison of the sum-rates (left) and 5\textsuperscript{th} percentile rates (right) achieved by models trained using $p+1\in\{2,4,6\}$ transmit power levels on networks with 4 APs and various numbers of UEs.}
\label{fig:self_4APs_diff_power_levels}
\end{figure*}

The results presented in Section~\ref{sec:results} were obtained using $p+1=2$ transmit power levels only. To ensure that our approach is applicable to the more general case of $p>1$ positive power levels, we trained and tested our algorithm with $N=4$ APs, using $p+1=4$ and $p+1=6$ power levels. The results are plotted in Figure~\ref{fig:self_4APs_diff_power_levels}. We observe that increasing the number of power levels from $p+1=2$ to $p+1=4$ produces a modest increase in sum-rate and that the performance of $p+1=4$ and $p+1=6$ power levels is nearly identical, confirming the fact that binary power control is already optimal in many scenarios of interest~\cite{gjendemsjo2008binary}.

\subsection{Enhanced Training Procedure}

One of the unique challenges of training agents to perform scheduling tasks in a multi-node wireless environment is that we can only simulate individual snapshots of the environment and explore a limited subset of the state space. In our training procedure, we fixed the parameters governing our environment (size of the deployment area, number of APs and UEs, minimum distance constraints, maximum transmit powers, etc.) and selected AP and UE locations randomly at each environment reset. We utilized multiple parallel environments to ensure that training batches contained experiences from different snapshots. This approach, along with a carefully-selected learning rate schedule and a sufficiently long training period, allowed our agents to achieve the performance that we reported in Section~\ref{sec:results}. An interesting question remains whether a more deliberate selection of training environments can accelerate training and/or result in better-performing agents. The fact that we can simulate only a limited number of environment realizations at a time provides an opportunity to guide the exploration and learning process. 

One potential direction is to control the density of the environments experienced during training, through systematically varying the parameters that govern the environment. Possible strategies could be to move from less dense to more dense deployments or vice versa, or to more carefully select training batches to always include experiences from a range of densities.

Another possible direction is to control the complexity of the environments on which the agent is being trained. Intuitively, some environment realizations are easy and some are more complex. For example, an episode where all of the UEs are very close to their associated APs, i.e., cell-center scenario, is easy because the optimal policy is for the APs to transmit at every scheduling interval. A scenario where all UEs are clustered around the borders between the coverage areas of adjacent APs, i.e., cell-edge scenario, is slightly more difficult because the optimal policy is for neighboring APs to coordinate not to transmit at the same time. Episodes in which users are distributed throughout the coverage areas of the APs are, however, much more complex because the optimal user scheduling and power control choices become completely non-trivial. Various approaches to curriculum learning could potentially be applied~\cite{hacohen2019power, seita2019zpd}. The main difficulty with this approach is determining a more granular measure of environment complexity and a procedure for generating environment realizations that exhibit the desired difficulty levels.

\subsection{Capturing Temporal Dynamics}
One of the main challenges faced by the scheduler is dealing with delayed observations available to the agent. We have shown that our agents are able to successfully cope with this problem, but an interesting question is whether we can include recurrent architectures in the agent's neural network to learn, predict, and leverage network dynamics based on the sequence of observations it receives over the course of an episode.

Two approaches are possible. The first is to include recurrent neural network (RNN) elements, such as long short-term memory (LSTM)~\cite{hochreiter1997long}, or attention mechanisms such as Transformers~\cite{vaswani2017attention}, at the inputs, with the goal of predicting the actual current observations based on the past history of delayed observations. 
A second approach would be to place the RNN/attention elements at the output similar to~\cite{reactor, r2d2}. One thing to note here is that for the system to learn the temporal dynamics of the environment, it must be exposed to the observations of each individual UE over time, implying that the approach of including observations from the top-3 UEs, sorted by their PF ratios, will likely not work and observations from all UEs must be included in an unsorted order. This is challenging due to the variable number of UEs associated to each AP, and we leave this for future work.

\section{Concluding Remarks}\label{sec:conc}

We introduced a distributed multi-agent deep reinforcement learning framework for performing joint user selection and power control  decisions in a dense multi-AP wireless network. Our framework takes into account real-world measurement, feedback, and backhaul communication delays and is scalable with respect to the size and density of the wireless network. We show, through simulation results, that our approach, despite being executed in a distributed fashion, achieves a tradeoff between sum-rate and 5\textsuperscript{th} percentile rate, which is close to that of a centralized information-theoretic scheduling algorithm. We also show that our algorithm is robust to variations in the density of the wireless network and maintains its performance gains even if the number of APs and/or UEs change(s) in the network during deployment.

Our multi-agent deep reinforcement learning framework demonstrates a practical approach for performing joint optimization of radio resource management decisions in real-world networks, taking into account measurement and communication delays and bounding the number of neural network inputs and outputs to reduce the computational complexity. Future research directions are to include additional resource management decisions, such as link adaptation, and to verify that an agent trained in simulation will perform as expected in a real network~\cite{peng2018sim}.

We believe that for the foreseeable future, learning-based resource management techniques, such as the one proposed in this paper, need to be augmented with expert policies that rely on optimization- and information-theoretical backbones, and cannot be viewed as an absolute replacement. Indeed, in the cases where the underlying assumptions are valid, theoretical frameworks can be used to manage the radio resources, while in the scenarios where those assumptions do not hold, learning-based algorithms can take over the resource management decisions.

\bibliographystyle{IEEEtran}
\bibliography{references}

\end{document}